\documentclass[10pt,journal,compsoc]{IEEEtran}
\usepackage{graphicx}
\usepackage{amsmath,amssymb}
\usepackage{color}
\usepackage{mathrsfs}
\usepackage{algorithm}
\usepackage{algpseudocode}
\usepackage{multirow}
\usepackage{caption}
\usepackage{subcaption}
\usepackage{diagbox}
\usepackage{fancyhdr}
\usepackage{pdfpages}

\newcommand{\etal}{\textit{et al}. }
\newcommand{\ie}{\textit{i}.\textit{e}. }
\newcommand{\eg}{\textit{e}.\textit{g}. }

\usepackage{xcolor,colortbl}
\definecolor{Gray}{gray}{0.85}
\definecolor{LightCyan}{rgb}{0.88,1,1}
\newcolumntype{a}{>{\columncolor{Gray}}c}

\ifCLASSOPTIONcompsoc
  \usepackage[nocompress]{cite}
\else
  \usepackage{cite}
\fi
\ifCLASSINFOpdf
\else
\fi
\begin{document}
%
\title{ZoomNAS: Searching for Whole-body Human Pose Estimation in the Wild}
%
%
%
%

\author{Lumin Xu, Sheng Jin, Wentao Liu, Chen Qian, Wanli Ouyang,~\IEEEmembership{Senior Member,~IEEE,} Ping Luo,~\IEEEmembership{Member,~IEEE,} Xiaogang Wang,~\IEEEmembership{Senior Member,~IEEE}
\IEEEcompsocitemizethanks{\IEEEcompsocthanksitem Lumin Xu and Xiaogang Wang are with The Chinese University of Hong Kong, Hong Kong SAR, China.\protect\\
E-mail: \{luminxu@link, xgwang@ee\}.cuhk.edu.hk

\IEEEcompsocthanksitem Sheng Jin and Ping Luo are with The University of Hong Kong, Hong Kong SAR, China.\protect\\
E-mail: \{js20@connect, pluo@cs\}.hku.hk

\IEEEcompsocthanksitem Wentao Liu and Chen Qian are with SenseTime Research and Tetras.AI.\protect\\
E-mail: \{liuwentao, qianchen\}@tetras.ai

\IEEEcompsocthanksitem Wanli Ouyang is with The University of Sydney and Shanghai AI Laboratory.\protect\\
E-mail: wanli.ouyang@sydney.edu.au
} 
\thanks{
Digital Object Identifier no. 10.1109/TPAMI.2022.3197352
}
}

%
%

\markboth{IEEE TRANSACTIONS ON PATTERN ANALYSIS AND MACHINE INTELLIGENCE}%
{Shell \MakeLowercase{\textit{et al.}}: Bare Demo of IEEEtran.cls for Computer Society Journals}
%



\IEEEtitleabstractindextext{%
\begin{abstract}
This paper investigates the task of 2D whole-body human pose estimation, which aims to localize dense landmarks on the entire human body including body, feet, face, and hands. We propose a single-network approach, termed ZoomNet, to take into account the hierarchical structure of the full human body and solve the scale variation of different body parts. We further propose a neural architecture search framework, termed ZoomNAS, to promote both the accuracy and efficiency of whole-body pose estimation. ZoomNAS jointly searches the model architecture and the connections between different sub-modules, and automatically allocates computational complexity for searched sub-modules. To train and evaluate ZoomNAS, we introduce the first large-scale 2D human whole-body dataset, namely COCO-WholeBody V1.0, which annotates 133 keypoints for in-the-wild images. Extensive experiments demonstrate the effectiveness of ZoomNAS and the significance of COCO-WholeBody V1.0.
\end{abstract}

\begin{IEEEkeywords}
Whole-body human pose estimation, neural architecture search, in-the-wild dataset
\end{IEEEkeywords}}

\maketitle

\thispagestyle{fancy}
\fancyhf{}
\renewcommand{\headrulewidth}{0pt}
\fancyfoot[C]{\scriptsize © 2022 IEEE. Personal use of this material is permitted. Permission from IEEE must be obtained for all other uses, in any current or future media, including reprinting/republishing this material for advertising or promotional purposes, creating new collective works, for resale or redistribution to servers or lists, or reuse of any copyrighted component of this work in other works.}

\IEEEdisplaynontitleabstractindextext

%
\IEEEpeerreviewmaketitle

\IEEEraisesectionheading{\section{Introduction}\label{sec:introduction}}

%
%
%
%

\IEEEPARstart{W}{hole-body}	human pose estimation aims at predicting keypoint localization of body, foot, face and hand joints in an image. This task is an extension and combination of four research related topics, \ie body/foot/face/hand keypoint estimation, which have been widely explored in the literature independently. With the development of computer vision applications, such as virtual reality, augmented reality and action recognition, whole-body pose estimation is attracting attention as it provides detailed information of all the human body parts. 

There are two existing approaches~\cite{cao2018openpose, hidalgo2019single} for 2D whole-body human pose estimation, which are based on deep convolutional neural networks.

OpenPose~\cite{cao2018openpose} applies multiple neural networks to estimate keypoints of different human parts. It first detects body and foot keypoints, then applies extra independent models for face and hand pose estimation. 
Since OpenPose uses multiple independent neural networks, it is not end-to-end trainable, resulting in sub-optimal whole-body keypoint localization results. 
It is desirable to have an end-to-end deep model to integrate multiple neural networks for different human parts.

Single-Network (SN)~\cite{hidalgo2019single} is such a design. Specifically, SN proposes a bottom-up end-to-end model to estimate whole-body keypoints simultaneously. However, SN does not take into account the scale variance of different person instances and different body parts, which hampers its effectiveness in localizing small-scale human body parts. In this paper, we aim to design an end-to-end model that can handle the scale variance of people and body parts.

\begin{figure}[tb]
	\centering
	\includegraphics[width=0.48\textwidth]{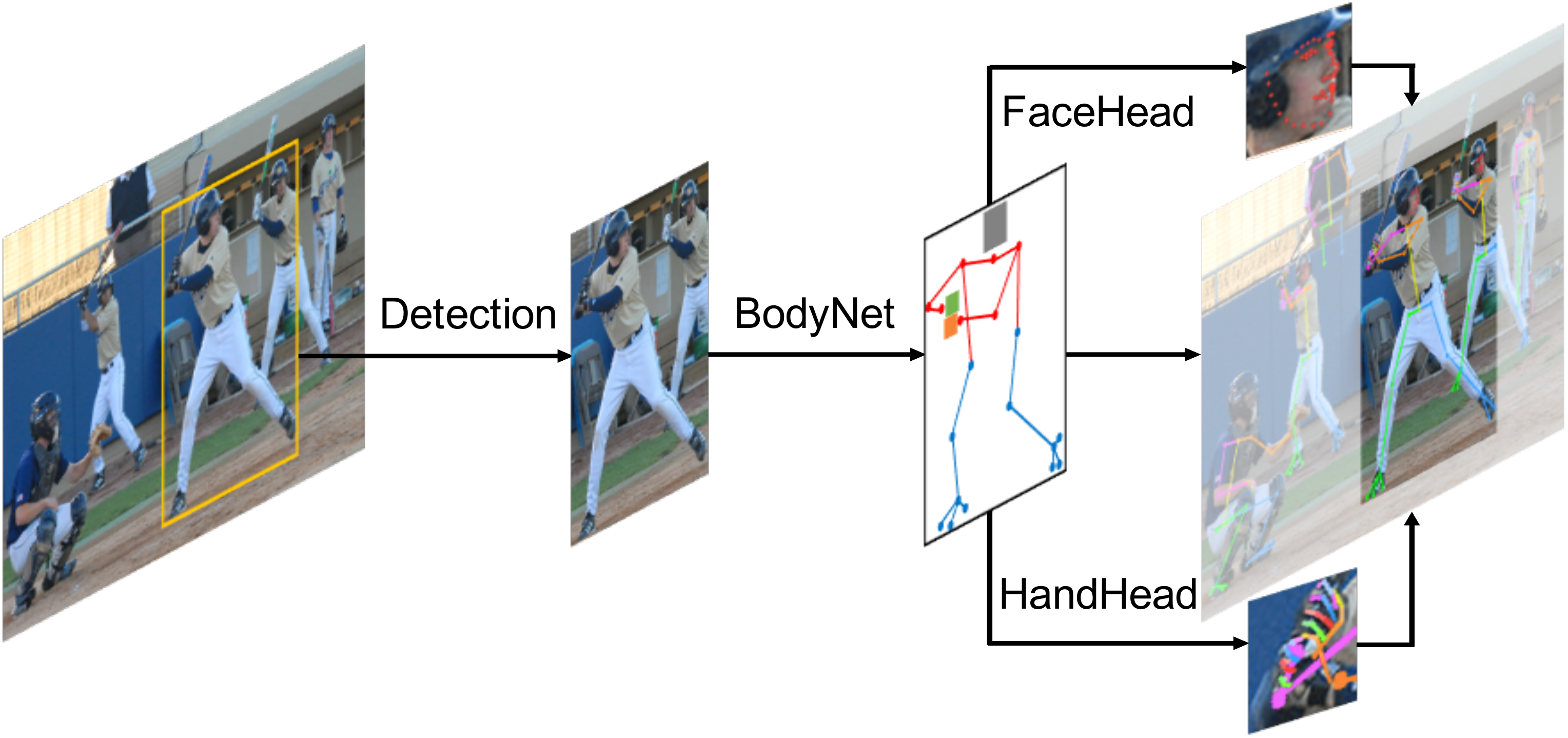}
	\caption{The pipeline of ZoomNet. ZoomNet adopts the top-down paradigm that first detects human instances and then estimates keypoints. A single neural network consisting of three modules, \ie BodyNet, FaceHead, and HandHead, is proposed to localize whole-body keypoints for each person. BodyNet predicts body/foot keypoints and face/hand boxes. FaceHead/HandHead zoom in to the face/hand areas and estimate face/hand keypoints at a higher resolution. This design handles the scale variance of different body parts of a person instance.}
	\label{fig:pipeline}
\end{figure}

This paper proposes the first end-to-end top-down whole-body pose estimation method, termed ZoomNet, as shown in Fig.~\ref{fig:pipeline}.
Following the top-down paradigm, our proposed ZoomNet first detects the individuals and then localizes the keypoints for each person. A multi-head neural network is designed for whole-body pose estimation, which consists of BodyNet and FaceHead/HandHead in a single network. The proposed single-network approach allows end-to-end training that jointly optimizes the whole-body keypoints. BodyNet estimates the keypoints of body/foot and the bounding boxes of face/hand. FaceHead/HandHead estimate the locations of face/hand keypoints.
ZoomNet handles the scale variance at two levels, \ie human instance level and face/hand level.
At the human instance level, as the detected persons are resized to the same scale, ZoomNet well handles the scale variance of different human instances. The body parts, except face/hand keypoints, are located at the human instance level. 
At the face/hand level, to get more detailed information of  face/hand, the features for estimating hand/face keypoints are further rescaled to higher resolution.
Specifically, FaceHead/HandHead take the features extracted by BodyNet as the input and zoom in to obtain higher resolution features for localizing the face/hand keypoints. 

However, manual trial-and-error is still required when designing the neural architectures of BodyNet/FaceHead/ HandHead and unifying these sub-modules into a single network, which is tedious and inefficient. To address the issue, we propose a neural architecture search (NAS) framework, termed ZoomNAS, to give a unified solution. As directly adopting the existing neural architectures for BodyNet/FaceHead/HandHead makes the whole-body model extremely heavy, we carefully design the search spaces of depth, channel, group number and input resolution, and search for the best configurations.
Moreover, FaceHead/HandHead localize face/hand keypoints using the features extracted by BodyNet. Generally, high-level features offer global contextual information, while low-level features contain detailed appearance information. Additionally, larger input feature areas can ensure the inclusion of target parts, but may cause inaccurate keypoint localization. As a result, we also search the connections between BodyNet and FaceHead/HandHead, \ie the choice of feature stage and the size of cropped area. To achieve the optimal trade-off between accuracy and efficiency, we train and search ZoomNAS (including BodyNet/FaceHead/HandHead) as a whole under the overall Flops constraint. The computational complexity is automatically allocated among these sub-modules. 
Experiments show that ZoomNAS significantly outperforms the prior arts on whole-body pose estimation with lower computational complexity.

These deep neural networks for whole-body pose estimation benefit from a large number of high-quality training data. The previous works utilize multiple datasets, including body~\cite{lin2014microsoft}/foot~\cite{cao2018openpose}/face~\cite{Gross2010Image,phillips2005overview,sagonas2013300}/hand~\cite{simon2017hand} keypoint datasets, to train models for whole-body pose estimation. However, the variations between these datasets, such as collection environment and data scales, introduce biases to neural network learning and hamper the performance from considering the task as a whole. To solve this problem, we propose COCO-WholeBody V1.0 dataset, the first large-scale benchmark for 2D whole-body pose estimation. COCO-WholeBody V1.0 extends COCO~\cite{lin2014microsoft} dataset with 133 dense landmarks including 68 on face, 42 on hands, and 23 on body and feet. Two subsets of COCO-WholeBody V1.0, WholeBody-Face (WBF) and WholeBody-Hand (WBH), are also introduced for face/hand pose estimation. They provide in-the-wild face/hand keypoint annotations and show strong generalization capacity by cross-dataset evaluation. 

The main contributions of our work are three-folds. 
\begin{itemize}
\item We introduce the first top-down single-network method for 2D whole-body human pose estimation, termed ZoomNet. 
It consists of BodyNet and FaceHead/HandHead, and localizes body/face/hand keypoints according to the hierarchical structure of the full human body, which addresses the scale variance of different human parts.
\item We apply neural architecture search for efficient neural network design. Our proposed ZoomNAS jointly searches the neural network architecture and connections between sub-modules. BodyNet/FaceHead/HandHead are searched simultaneously and the computational complexity is automatically allocated among them. ZoomNAS achieves a better trade-off between accuracy and efficiency, and outperforms the prior art, ZoomNet, by 2.4\% mAP with 37\% lower computation complexity. 
\item The first large-scale whole-body pose estimation dataset, termed COCO-WholeBody V1.0, is proposed to boost the related research. We annotate 133 keypoints (17 for body, 6 for feet, 68 for face, and 42 for hands) for more than 200K instances. We also introduce two subsets, WholeBody-Face (WBF) and WholeBody-Hand (WBH) for in-the-wild face/hand keypoint localization. Cross-dataset evaluations validate the powerful generalization ability of COCO-WholeBody V1.0.
\end{itemize}

A preliminary version of our work is published in~\cite{jin2020whole}. The main difference with~\cite{jin2020whole} is summarized as follows. 
1) We propose a neural architecture search (NAS) framework, termed ZoomNAS, to discover efficient networks for whole-body pose estimation based on ZoomNet. ZoomNAS designs special search spaces for the multi-head network and conducts effective computation allocation among the sub-modules. The effectiveness of the NAS approach is validated by experiments. 2) We further improve the annotations of our proposed COCO-WholeBody dataset from V0.5~\cite{jin2020whole} to V1.0. Both quality and quantity of the dataset are promoted and the improvement is demonstrated by evaluating the baseline methods on the same validation set. We also construct two sub-datasets WholeBody-Face (WBF) and WholeBody-Hand (WBH) for in-the-wild face/hand keypoint localization. Benchmarking experiments and cross-dataset evaluations are conducted on these two datasets.

\section{Related Work} \label{sec:related_work}

\subsection{Pose Estimation Method}
\textbf{Body Keypoint Localization.} The task of human body pose estimation~\cite{ouyang2014multi,wei2016convolutional,yang2016end,newell2016stacked,chu2017multi,yang2017learning} has been widely explored in the literature. Recent works mainly focus on multi-person pose estimation and fall into two categories, \ie bottom-up and top-down approaches. Bottom-up approaches~\cite{cao2017realtime,newell2017associative,nie2017generative,papandreou2018personlab,pishchulin2016deepcut,Insafutdinov2016ArtTrack,Insafutdinov2016DeeperCut,Iqbal2016PoseTrack,jin2019multi} first detect all the keypoints of every person in images and then group them into individuals. Top-down methods~\cite{he2017mask,papandreou2017towards,fang2017rmpe,chen2018cascaded,xiao2018simple,sun2019deep,liu2018cascaded} first detect the bounding box of each person and then predict the human body keypoints in each box. By resizing and cropping, top-down approaches normalize the poses to approximately the same scale. Therefore, they are more robust to human-level scale variance and recent state-of-the-arts are obtained by top-down approaches. However, directly using the existing top-down methods for whole-body pose estimation will encounter the problem of scale variance of different body parts (body vs face/hand). To tackle this problem, we propose ZoomNAS, a single-network top-down approach that zooms in to the hand/face regions and predicts the hand/face keypoints using higher feature resolution for accurate localization. 

\textbf{Face Keypoint Localization.}
Face keypoint localization, also referred to as facial landmark detection or face alignment, has been intensively studied in the literature. Recent deep learning based approaches can be divided into coordinate regression methods and heatmap regression methods. Coordinate regression methods~\cite{sun2013deep,zhang2015learning,trigeorgis2016mnemonic,feng2018wing} directly map the input image to the keypoint x-y coordinates. These models explicitly infer keypoint coordinates without any decoding. However, they do not perform as well as heatmap regression methods. Heatmap regression methods~\cite{yang2017stacked,deng2019joint,wu2018look,wang2020deep} generate heatmaps to represent the probability of the keypoint location. Cascaded networks~\cite{yang2017stacked,deng2019joint}, boundary maps~\cite{wang2019adaptive,wu2018look}, and uncertainty estimation~\cite{kumar2020luvli} are adopted to improve localization accuracy.

\textbf{Hand Keypoint Localization.} 
Most existing works on hand keypoint localization focus on 3D estimation. These works rely on auxiliary information such as depth information~\cite{oikonomidis2012tracking,sharp2015accurate,sridhar2015fast}, multi-view information~\cite{guan2006multi,neverova2014multi,simon2017hand} or 3D shape models~\cite{bouritsas2019neural,baek2019pushing,boukhayma20193d} with strong priors (\eg MANO~\cite{ranjan2018generating}). State-of-the-art models mainly adopt the two-stage framework~\cite{cai2018weakly,li2020exploiting,doosti2020hope,zimmermann2017learning,chen2020nonparametric}, which first regresses 2D keypoint locations then lifts 2D keypoints to 3D. For 2D keypoint localization, they~\cite{cai2018weakly,li2020exploiting} commonly adopt encoder-decoder architectures (\eg CPM~\cite{wei2016convolutional} or hourglass network~\cite{newell2016stacked}), which are initially developed for human pose estimation. A combination of weak 2D supervision on real images and full 3D joint supervision on synthetic images are leveraged to improve the generalization ability. 

\textbf{Foot Keypoint Localization.} 
The works on foot keypoint localization are very limited up to now, due to limited datasets. Cao \emph{et al.}~\cite{cao2018openpose} propose a bottom-up method to jointly learn foot and body keypoints. The joint training does not hurt the performance of body keypoint localization.

\textbf{Whole-Body Pose Estimation.} To the best of our knowledge, there are only two existing works targeting 2D whole-body pose estimation. OpenPose~\cite{cao2018openpose} applies multiple models to handle different kinds of keypoints. It predicts the keypoint localization in an image as well as the part affinity fields (PAFs) for keypoint grouping. OpenPose consists of three independent neural networks, including body pose network, hand keypoint network, and face landmark network. The body pose network estimates the body and foot keypoints and the corresponding PAFs. The hand keypoint network localizes the hand keypoints in the hand bonding boxes generated with the arm keypoints, and the face landmark network infers the face keypoints around the rough facial locations. 
Since OpenPose relies on multiple networks, it is hard to train and suffers from increased runtime and computational complexity. Unlike OpenPose, our proposed ZoomNAS is a single-network method as it integrates five previously separated models (human body pose estimator, hand/face box detectors, and hand/face pose estimators) into a single network with shared features. Recently, Hidalgo \etal proposes an elegant method SN~\cite{hidalgo2019single} for bottom-up whole-body keypoint estimation. SN is based on PAF~\cite{cao2017realtime} which predicts the keypoint heatmaps for detection and part affinity maps for grouping. Since there is no such dataset with whole-body annotations, they use a set of different datasets and carefully design the sampling rules to train the model. However, bottom-up approaches cannot handle the scale variation problem well and would have difficulty in detecting face and hand keypoints accurately. In comparison, our ZoomNAS is a top-down approach that well handles the scale variance problem. It also takes into account the inherent hierarchical structure of the full human body to solve the scale variation of different parts in the same person. Also, automatic network design via neural architecture search for whole-body pose estimation is not investigated in OpenPose or SN.

Recently, the task of recovering 3D human mesh~\cite{kanazawa2018end,lin2021end,lin2021mesh} has been extended to the task of monocular 3D whole-body capture~\cite{joo2018total,romero2017embodied,xiang2019monocular}. Romero \etal\cite{romero2017embodied} proposes a generative 3D model to express body and hands. Xiang \etal\cite{xiang2019monocular} introduces a 3D deformable human model to reconstruct whole-body pose and Joo \etal\cite{joo2018total} presents Adam which encompasses the expressive power for body, hand, and facial expression. These methods still rely on OpenPose~\cite{cao2018openpose} to localize 2D body keypoints in images. 

\begin{table*}
	\caption{Overview of some popular public datasets for 2D keypoint estimation. ``Kpt'' stands for keypoints, and ``\#Kpt'' means the annotated number. ``Wild'' denotes whether the dataset is collected in-the-wild. ``*'' means the head box.}
	\begin{center}
		\begin{tabular}{c|c|c|c|ccc|ccc|c}
			\hline
			DataSet & Images & \#Kpt & Wild &  Body   &    Hand    &    Face   &   Body   &    Hand    &   Face & Total \\ 
			&        &      &        & Box     &    Box     &    Box    &   Kpt    &    
			Kpt     &    Kpt & Instances \\\hline\hline
			\textit{MPII}~\cite{andriluka20142d}      & 25K & 16 & \checkmark &  \checkmark &   & * & \checkmark &    & & 40K \\
			\textit{MPII-TRB}~\cite{duan2019trb}      & 25K & 40 & \checkmark &  \checkmark &   & * & \checkmark &    & & 40K \\
			\textit{CrowdPose}~\cite{li2019crowdpose} & 20K & 14 & \checkmark  & \checkmark &   & & \checkmark & &   & 80K  \\ 
			\textit{PoseTrack}~\cite{andriluka2018posetrack} & 23K & 15 & \checkmark  & \checkmark &   & & \checkmark & &   & 150K  \\ 
			\textit{AI Challenger}~\cite{wu2017ai}            &  300K & 14 & \checkmark & \checkmark  &   &  & \checkmark &    &  & 700K  \\
			\textit{COCO}~\cite{lin2014microsoft}    & 200K & 17  & \checkmark & \checkmark  &   & * & \checkmark &    &  & 250K  \\\hline
			\textit{OneHand10K}~\cite{Yangang2018Mask}            & 10K & 21 &\checkmark  &    & \checkmark & &  & \checkmark & & - \\
			\textit{SynthHand}~\cite{mueller2017real}   & 63K &  21 &  &    & \checkmark & &  & \checkmark & & - \\
			\textit{RHD}~\cite{zb2017hand}            & 41K &  21 &  &    & \checkmark & &  & \checkmark & & - \\
			\textit{FreiHand}~\cite{Freihand2019}     & 130K & 21 &  &    &  & &  & \checkmark & & - \\
			\textit{MHP}~\cite{gomez2017large}        & 80K & 21  &  &    & \checkmark & &  & \checkmark & & - \\
			\textit{GANerated}~\cite{mueller2018ganerated}  & 330K & 21 &   &   &    & &  & \checkmark & & - \\
			\textit{Panoptic}~\cite{simon2017hand}    & 15K & 21 &  &   &  \checkmark & &  & \checkmark & & - \\ \hline
			\textit{WFLW}~\cite{wu2018look}    & 10K  &  98 & \checkmark  &   &   & \checkmark & &    & \checkmark & - \\
			\textit{AFLW}~\cite{koestinger2011annotated}    & 25K  & 19 & \checkmark  &   &   & \checkmark & &    & \checkmark & - \\
			\textit{COFW}~\cite{Burgos2013Robust}    & 1852 & 29 & \checkmark  &   &   & \checkmark & &    & \checkmark & - \\
			\textit{300W}~\cite{sagonas2013300}            & 3837 & 68 & \checkmark &   &   & \checkmark & & & \checkmark & -      \\ \hline
			COCO-WholeBody & 200K & 133 & \checkmark  & \checkmark & \checkmark   &\checkmark  &  \checkmark & \checkmark & \checkmark & 250K  \\ \hline	
		\end{tabular}
	\end{center}
	\label{tab:dataset}
\end{table*}

\subsection{Neural Architecture Search}

\textbf{NAS for image classification.}
Neural architecture search (NAS) is proposed to search for the appropriate neural network architecture. Most NAS methods focus on the image classification task. Early NAS approaches~\cite{liu2018progressive,real2019regularized,tan2019mnasnet,zoph2016neural,zoph2018learning} search for the optimal architecture in a nested manner. They sample a great number of architecture candidates and train each architecture from scratch, which is extremely computationally expensive. Recent NAS approaches~\cite{cai2020once,cai2018proxylessnas,li2020improving,liang2019computation,liu2018darts,liu2021inception,liu2020block,wu2019fbnet,yu2020bignas,zhou2020econas} consider all the architectures in the search spaces as a whole. They use a weight sharing strategy, in which  a single super-network is trained. Each architecture candidate is a sub-network of the super-network (a sub-network has a subset of parameters and features of the super-network) and the weights of the candidate architectures are from the trained super-network, \ie the weights of candidate architectures are shared. Our work also follows the weight sharing paradigm. We train the super-network once, then sample and evaluate various sub-networks to discover the optimal architecture.

\textbf{NAS for human pose estimation.}
Recently, neural architecture search is applied to the human pose estimation task and promotes the success of efficient keypoint localization. PoseNFS~\cite{yang2019pose} proposes to learn specific micro and macro neural architectures for different human parts utilizing the prior knowledge of human structure. AutoPose~\cite{gong2020autopose} searches for multi-scale fusion and high-resolution representations by applying a reinforcement learning based bi-level optimization method. EfficientPose~\cite{zhang2020efficientpose} implements a differentiable neural architecture search method for the efficient backbone and a spatial information correction module for the efficient head. PoseNAS~\cite{bao2020pose} adopts a Fusion-and-Enhancement manner with scale-adaptive fusion cells for pose decoder, and end-to-end searches the pose encoder and pose decoder simultaneously. ViPNAS~\cite{xu2021vipnas} proposes a spatial-temporal neural architecture search framework for efficient video pose estimation. 

Our ZoomNAS is different from these NAS methods for human pose estimation on two aspects. First, neural architecture search method has not been studied for the more challenging task --- whole-body pose estimation.
Secondly, ZoomNAS is specifically designed for this task. It is shown in~\cite{radosavovic2019network} that search spaces are critical for the success of NAS. We propose the new search spaces to search the connections between the sub-modules for body, face, and hand, which is a special problem for whole-body pose estimation and has not been explored in the previous works. Also, ZoomNAS automatically allocates the computational resources among the sub-modules for body, face, and hand.

\subsection{Pose Estimation Benchmark}
There are several existing datasets for 2D pose estimation. However, they separately annotate the keypoint localization of body  \cite{eichner2010we,andriluka20142d,lin2014microsoft,wu2017ai,andriluka2018posetrack}, hand \cite{yuan2017bighand2,tompson2014real,gomez2017large,mueller2018ganerated,simon2017hand} or face \cite{koestinger2011annotated,sagonas2013300,belhumeur2013localizing,zhu2012face,le2012interactive,messer1999xm2vtsdb} in images or videos. We summarize these datasets in Table~\ref{tab:dataset} and give a brief introduction in this section.

\textbf{Multi-Person Body Pose Dataset.}
Multi-person pose estimation has achieved significant progress in the past few years, and there are a good few popular datasets. Most of these datasets~\cite{andriluka20142d,lin2014microsoft,wu2017ai,li2019crowdpose,duan2019trb} annotate body keypoints for images in the wild. PoseTrack~\cite{andriluka2018posetrack} provides dense annotations of video sequences with 15 body keypoints. COCO~\cite{lin2014microsoft} is one of the most popular datasets for body keypoint localization. It offers 17-keypoint body annotations in challenging, uncontrolled conditions. Our COCO-WholeBody V1.0 dataset is an extension of COCO, with densely annotated 133 body/foot/face/hand keypoints. The task of whole-body pose estimation is much more challenging than the traditional body pose estimation, due to 1) higher localization accuracy required for face/hand and 2) scale variance between body and face/hand.

\textbf{RGB-based Hand Keypoint Dataset.} Most existing RGB-based hand keypoint datasets are either synthetic~\cite{zb2017hand,mueller2018ganerated} or captured in the lab environment~\cite{Freihand2019,gomez2017large,simon2017hand}. For example, \emph{Panoptic}~\cite{simon2017hand} is a well-known hand keypoint dataset, recorded in the CMU's Panoptic studio with multiview dome settings. However, it is limited to a controlled laboratory environment with a simple background. OneHand10K~\cite{Yangang2018Mask} is a recent in-the-wild 2D hand keypoint dataset. However, its size is limited. Our WheleBody-Hand (WBH) dataset is complementary to these RGB-based hand keypoint datasets. It contains about 80K 21-keypoint labeled hands that are captured in the wild. To the best of our knowledge, WBH is the largest in-the-wild dataset for 2D RGB-based hand keypoint estimation. It is very challenging, due to occlusion, hand-hand interaction, hand-object interaction, motion blur, and small scales.

\textbf{Face Keypoint Dataset.} Face keypoint datasets~\cite{Burgos2013Robust,koestinger2011annotated,wu2018look,sagonas2013300} play a crucial role for the development of facial landmark detection a.k.a. face alignment. Among them, \emph{300W}\cite{sagonas2013300} is the most popular. It is a combination of LFPW\cite{belhumeur2013localizing}, AFW\cite{zhu2012face}, HELEN\cite{le2012interactive}, XM2VTS\cite{messer1999xm2vtsdb} with 68 landmarks annotated for each face image. Our proposed WholeBody-Face (WBF) dataset follows the same annotation settings as 300W and more than 50K 68-keypoint annotations are provided for face images. Compared to 300W, WBF contains much more annotations and is more challenging as it contains more blurry and small-scale facial images. 

To the best of our knowledge, our proposed COCO-WholeBody V1.0 is the first large-scale dataset for 2D whole-body human pose estimation. Different from the above datasets that target at body/face/hand keypoint localization separately, COCO-WholeBody annotates 133 body/foot/face/hand keypoints for multiple persons from in-the-wild images, which offers the unified representation for all the human-related keypoints. This novel dataset allows the whole-body pose estimation methods to take into account the correlations between different human parts, and encourages the further exploration of this task and downstream applications.

There are datasets that generate the 3D annotations of the full human body, which are also related to our work. For example, DensePose~\cite{alp2018densepose} provides a dense 3D surface-based representation for human shape. However, since the keypoints in DensePose are uniformly sampled, they lack specific joint articulation information and details of face/hands are missing.

\section{ZoomNet: Whole-body Pose Estimation}

In this section, we introduce the whole-body pose estimation pipeline of our proposed ZoomNet as shown in Fig.~\ref{fig:pipeline}. ZoomNet adopts the top-down paradigm, which first detects persons and then predicts keypoints for each person. Following the common setting ~\cite{xiao2018simple,sun2019deep}, we use the off-the-shelf FasterRCNN~\cite{renNIPS15fasterrcnn} detector to detect persons. For each detected person, ZoomNet hierarchically localizes the whole-body keypoints based on the full body structure. ZoomNet applies a multi-head architecture, which consists of convolutional neural networks including three sub-modules, BodyNet, FaceHead, and HandHead. BodyNet takes the cropped image of the detected person as the input and estimates the body/foot keypoints and face/hand bounding boxes simultaneously. FaceHead/HandHead predict face/hand keypoints based on the features of the face/hand areas from BodyNet.

\subsection{BodyNet}
Given a cropped and resized image of a detected person instance, BodyNet localizes the keypoints of body/foot and the bounding boxes of face/hand at the same time. 

BodyNet localizes the bounding box by 5 keypoints (4 corner points and 1 center point), which is inspired by CornerNet~\cite{law2018cornernet}, a one-stage detector that represents the object with keypoint pairs. For each person, the box keypoints of the face box, the left hand box, and the right hand box are predicted together with the body/foot keypoints. During inference, each box is obtained by the minimum bounding rectangle of the box keypoints.

We use 2D heatmaps to encode both the human keypoints and the box keypoints. BodyNet outputs 38-channel heatmaps for each person (17 for body, 6 for feet, and 15 for face/hand bounding boxes). HRNet~\cite{sun2019deep} is used as the architecture of BodyNet, which is a multi-resolution network and has shown the state-of-the-art performance on the human body pose estimation task.

\subsection{FaceHead and HandHead}
FaceHead/HandHead take the cropped features of predicted face/hand bounding boxes as the input. 
To provide detailed localization information for accurate face/hand keypoint estimation, the input features are extracted from the highest-resolution branch of BodyNet. RoIAlign~\cite{he2017mask} is applied for feature extraction, and visual features are resized to a higher resolution. HandHead and FaceHead predict the heatmaps of face/hand keypoints using the extracted features in parallel. The left hand and right hand are processed by the same HandHead. HRNetV2~\cite{wang2020deep}
is chosen as the base architecture of FaceHead and HandHead, which is proved to be effective for these sub-tasks.

As a brief summary, BodyNet detects the bounding boxes of face/hand and extracts the visual features, then FaceHead/HandHead zoom in to focus on the face/hand regions for accurate keypoint estimation. In this way, ZoomNet considers the hierarchical structure of the full human body and solves the problem of scale variance of different parts of a person instance. It extracts high-resolution features for hand/face keypoint estimation, and features with larger receptive fields for body keypoint estimation at the same time.

\subsection{Training Process}
\label{sec:train_zoomnet}

In the training process, we crop the person instances in the original images according to the ground truth human bounding boxes. Following the common setting of top-down pose estimation approaches~\cite{xiao2018simple,sun2019deep}, we train ZoomNet with data augmentation upon the human bounding boxes to improve the model generalization ability. We generate the ground truth heatmaps of body/foot keypoints and face/hand boxes to supervise BodyNet. Features extracted by BodyNet are cropped according to the predicted face/hand boxes, which serve as the input of FaceHead/HandHead. The ground truth heatmaps of face/hand keypoints are generated using the same boxes. Pixel-wise mean squared error (MSE) loss is applied to all the output heatmaps as follows,

\begin{equation}
    \mathcal{L}_{MSE} = \dfrac{1}{KHW} \sum_{k=1}^K \sum_{i=1}^H \sum_{j=1}^W (y_{k,i,j} - \hat{y}_{k,i,j})^2,
\label{eq:mse}
\end{equation}

\noindent where $K$ is the number of joints to be supervised, and $H$ and $W$ respectively denote the height and width of heatmaps. $y_{k,i,j}$ and $\hat{y}_{k,i,j}$ are the ground truth pixel intensity and the predicted pixel intensity in the corresponding position.

In our preliminary version~\cite{jin2020whole}, BodyNet/FaceHead/ HandHead are optimized step-by-step. This successive training strategy first trains BodyNet to provide reliable face/hand boxes and then optimizes FaceHead/HandHead in the corresponding regions, which lacks efficiency. Since ZoomNet is a single-network approach and is end-to-end trainable, this paper changes the step-by-step successive training in~\cite{jin2020whole} to joint training of BodyNet/FaceHead/HandHead and optimizes the whole-body keypoints at once. The overall training losses are as follows,

\begin{equation}
    \mathcal{L} =  \mathcal{L}_{body} +  \mathcal{L}_{face} +  \mathcal{L}_{hand},
\label{eq:loss}
\end{equation}

\noindent where $\mathcal{L}_{body}$, $\mathcal{L}_{face}$ and $\mathcal{L}_{hand}$ are the MSE losses of BodyNet, FaceHead, and HandHead respectively. $\mathcal{L}_{body}$ supervises the 38-channel outputs of BodyNet, including the keypoints of body/foot and the bounding box keypoints of face/hand. $\mathcal{L}_{face}$ and $\mathcal{L}_{hand}$ respectively supervises the 68 and 21 channel outputs of FaceHead and HandHead. 
Experimental results on the effectiveness and efficiency of joint training strategy are explored in Sec.~\ref{sec:effect of joint training}.

\section{ZoomNAS: Neural Architecture Search}

\begin{figure*}[tb]
	\centering
	\includegraphics[width=0.9\textwidth]{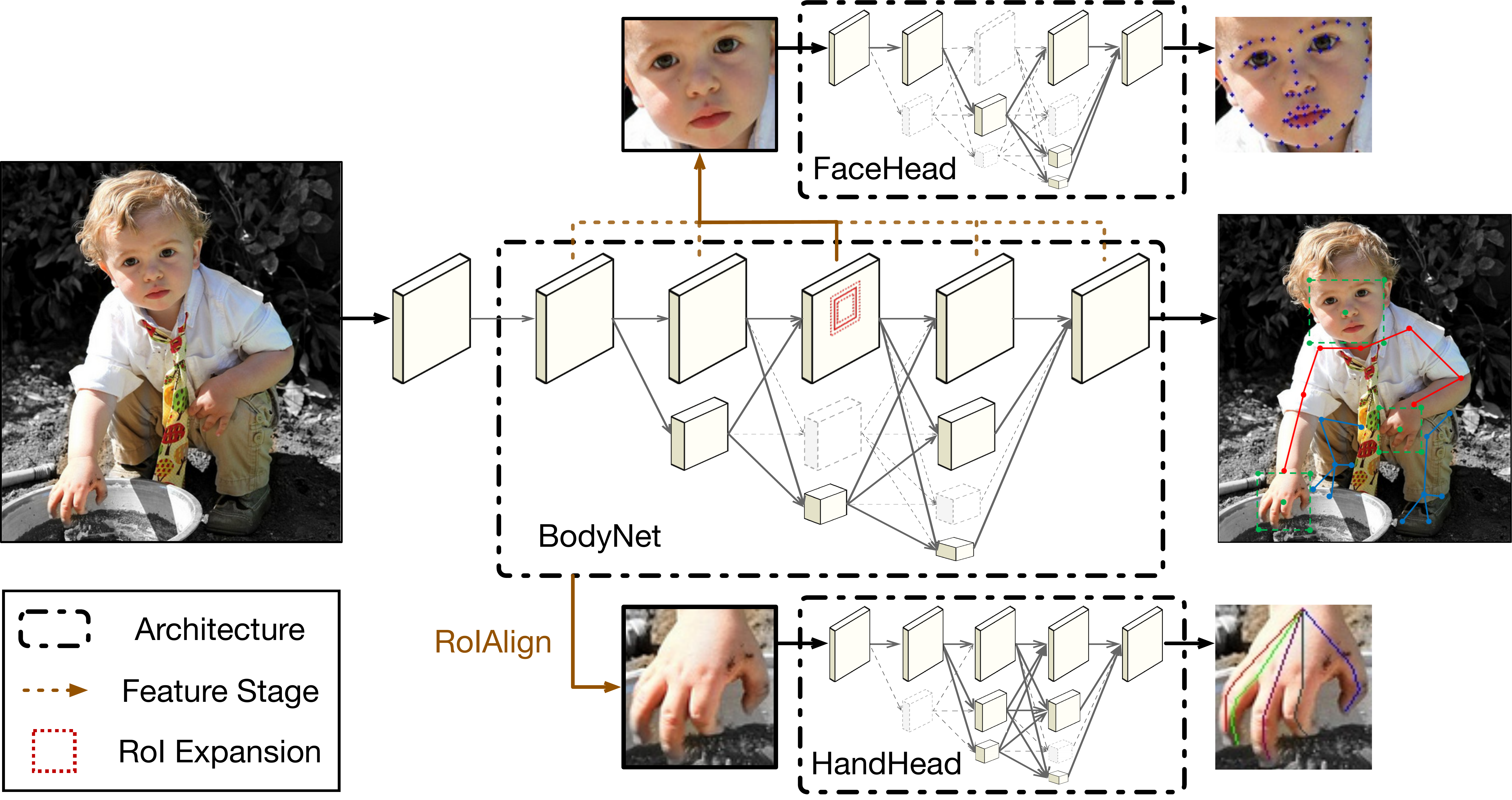}
	\caption{ZoomNAS applies neural architecture search to ZoomNet for efficient whole-body pose estimation. The architectural configurations of BodyNet/FaceHead/HandHead, including depth, channel, group number, and input resolution, are searched simultaneously. ZoomNAS also searches the connections between BodyNet and FaceHead/HandHead, \ie feature stage and RoI expansion. The computational complexity is allocated among different sub-modules automatically.}
	\label{fig:zoomnas}
\end{figure*}

\begin{table*}[tb]
	\begin{center}
	\caption{The search spaces of ZoomNAS. The discrete search space of each operator is denoted as [start, end; stride] or listed in \{ , ,  \dots, \}. Both height and width of images (or features) are used to represent the resolution.
	}
	 \scalebox{0.83}{
		\begin{tabular}{ccc|ccccccc}
			\hline
			 & \multicolumn{2}{c|}{\textbf{Connection}} & \multicolumn{7}{c}{\textbf{Architectural Configuration}} \\
			 & Feature Stage & RoI Expansion & Stage & Branch & Operator & Depth & Channel & Group Number & Resolution \\  \hline \hline
    		\multirow{12}{*}{BodyNet} & \multirow{12}{*}{-} & \multirow{12}{*}{-} &  &  & \multirow{2}{*}{Crop \& Resize} & \multirow{2}{*}{-} & \multirow{2}{*}{-} & \multirow{2}{*}{-} & $H$ = [256, 384; 32], \\ 
    		~ & ~ &  &  & ~ & ~ & ~ & ~ & ~ & $W = 3H/4$ \\ \cline{4-10}
    	    ~ & ~ & ~ & 0 & - & Convolution & [2, 2; 1] & $C_0$ = [16, 64; 16] & [1, 1; 1] & $H /4$, $W /4$ \\  \cline{4-10}
		    ~ & ~ & ~ & 1 & 1 & Bottleneck & [2, 4; 1] & $C$ = [16, 64; 16] & $\{2^0 C$, $2^{-1} C$, \dots, $2^{-6} C\}$ & $H/4$, $W/4$ \\  \cline{4-10}
		    ~ & ~ & ~ & \multirow{2}{*}{2} & 1 & \multirow{2}{*}{BasicBlock} & \multirow{2}{*}{[4, 4; 4]} & $C_1$ = [8, 32; 8] & $\{2^0 C_1 $, $2^{-1} C_1 $, \dots, $2^{-5} C_1 \}$ & $H/4$, $W/4$ \\ 
		    ~ & ~ & ~ & ~ & 2 & ~ & ~ & $C_2$ = [16, 64; 16] & $\{2^0 C_2 $, $2^{-1} C_2 $, \dots, $2^{-6} C_2 \}$ & $H/8$, $W/8$  \\  \cline{4-10}
		    ~ & ~ & ~ & \multirow{3}{*}{3} & 1 & \multirow{3}{*}{BasicBlock} & \multirow{3}{*}{[8, 16; 4]} & $C_1$ = [8, 32; 8] & $\{2^0 C_1 $, $2^{-1} C_1 $, \dots, $2^{-5} C_1 \}$ & $H/4$, $W/4$ \\
		    ~ & ~ & ~ & ~ & 2 & ~ & ~ & $C_2$ = [16, 64; 16] & $\{2^0 C_2 $, $2^{-1} C_2 $, \dots, $2^{-6} C_2 \}$ & $H/8$, $W/8$ \\
		    ~ & ~ & ~ & ~ & 3 & ~ & ~ & $C_3$ = [32, 128; 32] & $\{2^0 C_3$, $2^{-1} C_3$, \dots, $2^{-7} C_3\}$ & $H/16$, $W/16$ \\  \cline{4-10}
		    ~ & ~ & ~ & \multirow{4}{*}{4} & 1 & \multirow{4}{*}{BasicBlock} & \multirow{4}{*}{[8, 12; 4]} & $C_1$ = [8, 32; 8] & $\{2^0 C_1 $, $2^{-1} C_1 $, \dots, $2^{-5} C_1 \}$ & $H/4$, $W/4$ \\
		    ~ & ~ & ~ & ~ & 2 & ~ & ~ & $C_2$ = [16, 64; 16] & $\{2^0 C_2 $, $2^{-1} C_2 $, \dots, $2^{-6} C_2 \}$ & $H/8$, $W/8$ \\
		    ~ & ~ & ~ & ~ &  3 & ~ & ~ & $C_3$ = [32, 128; 32] & $\{2^0 C_3$, $2^{-1} C_3$, \dots, $2^{-7} C_3\}$ & $H/16$, $W/16$ \\
		    ~ & ~ & ~ & ~ & 4 & ~ & ~ & $C_4$ = [64, 256; 64] & $\{2^0 C_4$, $2^{-1} C_4$, \dots, $2^{-8} C_4\}$ & $H/32$, $W/32$ \\  \hline \hline
		    
		    \multirow{11}{*}{\shortstack{FaceHead/ \\ HandHead}} &  \multirow{11}{*}{[0, 1, 2, 3, 4]} &  \multirow{11}{*}{[1.0, 1.3; 0.1]} &  &  & \multirow{2}{*}{RoIAlign \& Resize} & \multirow{2}{*}{-} & \multirow{2}{*}{-} & \multirow{2}{*}{-} & $H$ = [32, 96; 16], \\
		    ~ & ~ &  &  & ~ & ~ & ~ & ~ & ~ & $W = H$ \\ \cline{4-10}
		    ~ & ~ & ~ & 1 & 1 & Bottleneck & [2, 4; 1] & $C$ = [16, 64; 16] & $\{2^0 C$, $2^{-1} C$, \dots, $2^{-6} C\}$ & $H$, $W$ \\  \cline{4-10}
		    ~ & ~ & ~ & \multirow{2}{*}{2} & 1 & \multirow{2}{*}{BasicBlock} & \multirow{2}{*}{[4, 4; 4]} & $C_1$ = [8, 32; 8] & $\{2^0 C_1$, $2^{-1} C_1$, \dots, $2^{-5} C_1 \}$ & $H$, $W$ \\  
		    ~ & ~ & ~ & ~ & 2 & ~ & ~ & $C_2$ = [16, 64; 16] & $\{2^0 C_2 $, $2^{-1} C_2 $, \dots, $2^{-6} C_2 \}$ & $H/2$, $W/2$  \\  \cline{4-10}
		    ~ & ~ & ~ & \multirow{3}{*}{3} & 1 & \multirow{3}{*}{BasicBlock} & \multirow{3}{*}{[8, 16; 4]} & $C_1$ = [8, 32; 8] & $\{2^0 C_1$, $2^{-1} C_1$, \dots, $2^{-5} C_1\}$ & $H$, $W$ \\
		    ~ & ~ & ~ & ~ & 2 & ~ & ~ & $C_2$ = [16, 64; 16] & $\{2^0 C_2 $, $2^{-1} C_2 $, \dots, $2^{-6} C_2 \}$ & $H/2$, $W /2$ \\
		    ~ & ~ & ~ & ~ & 3 & ~ & ~ & $C_3$ = [32, 128; 32] & $\{2^0 C_3$, $2^{-1} C_3$, \dots, $2^{-7} C_3\}$ & $H /4$, $W /4$ \\  \cline{4-10}
		    ~ & ~ & ~ & \multirow{4}{*}{4} & 1 & \multirow{4}{*}{BasicBlock} & \multirow{4}{*}{[8, 12; 4]} & $C_1$ = [8, 32; 8] & $\{2^0 C_1 $, $2^{-1} C_1 $, \dots, $2^{-5} C_1 \}$ & $H$, $W$ \\
		    ~ & ~ & ~ & ~ & 2 & ~ & ~ & $C_2$ = [16, 64; 16] & $\{2^0 C_2 $, $2^{-1} C_2 $, \dots, $2^{-6} C_2 \}$ & $H/2$, $W/2$ \\
		    ~ & ~ & ~ & ~ &  3 & ~ & ~ & $C_3$ = [32, 128; 32] & $\{2^0 C_3$, $2^{-1} C_3$, \dots, $2^{-7} C_3\}$ & $H/4$, $W/4$ \\
		    ~ & ~ & ~ & ~ & 4 & ~ & ~ & $C_4$=[64, 256; 64] & $\{2^0 C_4$, $2^{-1} C_4$, \dots, $2^{-8} C_4\}$ & $H/8$, $W/8$ \\  \hline
		\end{tabular}
	}
	\label{tab:search_space}
	\end{center}
\end{table*}

In this paper, we propose a neural architecture search framework based on ZoomNet, termed ZoomNAS, for a better trade-off between accuracy and efficiency as shown in Fig.~\ref{fig:zoomnas}. Motivated by~\cite{cai2020once,yu2020bignas}, we design the weight-shared super-network and search for the best-performing sub-network. The architectures of BodyNet/FaceHead/HandHead are searched on the search spaces of depth, channel, group number, and input resolution. The relationships between BodyNet and FaceHead/HandHead are also searched, including feature stage and RoI expansion. We consider the model as a whole and the computational complexity is automatically allocated among the searched sub-modules.

\subsection{Search Space}

As shown in Table~\ref{tab:search_space}, the search spaces of ZoomNAS consist of the detailed architectural configurations and connections between BodyNet and FaceHead/HandHead. The super-networks of BodyNet and FaceHead/HandHead are constructed based on HRNet~\cite{sun2019deep} and HRNetV2~\cite{wang2020deep} respectively. Network depth is searched at each stage. The channel number and the group number are searched at the branch level. The operators in a branch that maintains the feature resolution share the same number of output channels and groups. We search the height of input images or features for BodyNet/FaceHead/HandHead. The ratio of width and height is fixed as 3:4 for BodyNet, and 1:1 for FaceHead/HandHead. To search the connections of different sub-modules, the feature stage and RoI expansion are searched for FaceHead/HandHead to obtain the input features from BodyNet.

\subsubsection{Detailed Architectural Configurations}

The detailed architectural configurations of BodyNet, FaceHead and HandHead are searched, including depth, channel, group number, and input resolution.

\textbf{Elastic Depth}: The number of operators (\eg convolutional layer, Bottleneck~\cite{he2016deep}, and BasicBlock~\cite{he2016deep}) at each model stage. For a stage with $N$ operators in the super-network, we keep the first $D$ operators and skip the remaining $N-D$ operators, when the depth $D$ is selected.

\textbf{Elastic Channel}: The number of output channels of each operator. We keep the first $C$ output channels when the width $C$ is selected.

\textbf{Elastic Group Number}: The number of groups in the grouped convolutional layers~\cite{krizhevsky2017imagenet} of each operator. For a convolutional layer with $C$ output channels, the possible choices of group number $G$ are $G = 2^{-i} \times C$ where $i \in \{0, 1, 2, \dots \}$ and $G \in \mathbb{N_+}$.

\textbf{Input Resolution}: The resolution of input images or features. Different human parts prefer different input sizes for keypoint localization. We keep the aspect ratio of the inputs and  only search the height, which corresponds to the size of images/features. The input images or features are resized to the selected resolution and fed into the network.

\subsubsection{Connections of Sub-modules}
We propose to search the connections between BodyNet and FaceHead/HandHead since BodyNet provides the cropped features as the input of FaceHead/HandHead. We search the connections on two dimensions, \ie feature stage and RoI expansion, as follows: 

\textbf{Feature Stage}: The level of BodyNet features for FaceHead/HandHead. Generally, low-level features provide more details for localization, while high-level features contain more global context information. FaceHead and HandHead may prefer a certain level of features and the best feature stage is searched automatically. Considering HRNet~\cite{sun2019deep} backbone of BodyNet, features at each stage consist of several different resolution branches. As multi-scale fusion is applied between different resolution features at each stage, we choose features of one branch for simplicity. Empirically, we select the features from the highest resolution branch of BodyNet, because high-resolution features are critical for accurate face/hand keypoint estimation. 

\textbf{RoI Expansion}: The expansion ratio of face/hand regions to be cropped. As the face/hand boxes obtained by BodyNet are imperfect, the face/hand boxes may not cover the whole face/hand regions. Following the common setting of top-down methods, we enlarge the boxes for better performance. However, choosing an appropriate expansion ratio of region of interest (RoI) is important but laborious. Small boxes may miss some keypoints, while large boxes may include too much cluttered background and hamper accurate localization. We add it into our search spaces and search for the best RoI expansion. The features obtained by BodyNet are cropped according to the enlarged boxes, and are fed into FaceHead/HandHead as the input.

\begin{figure*}[tb]
	\centering
	\includegraphics[width=0.98\textwidth]{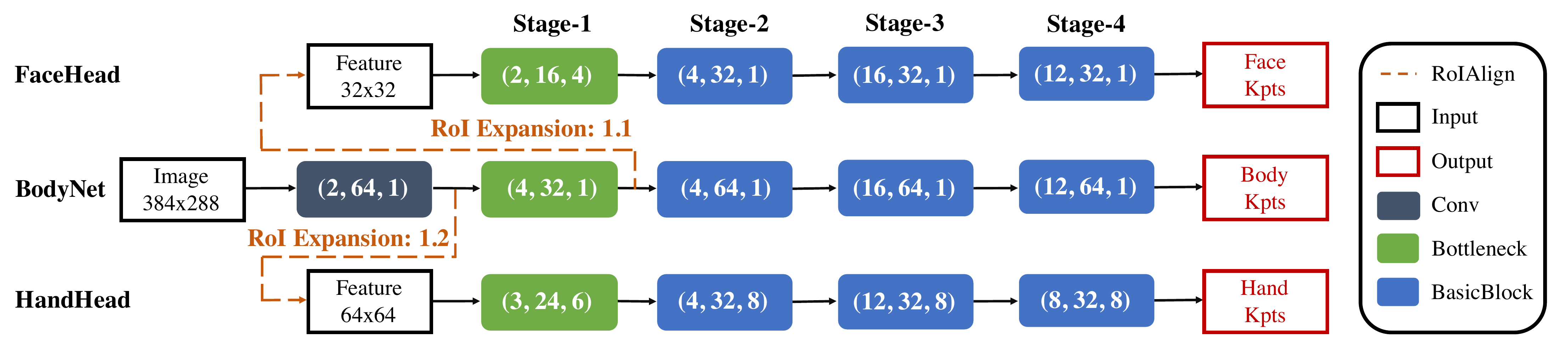}
	\caption{The discovered architecture of ZoomNAS. (Depth, Channel, Group) are listed for each stage. Only the branch with the highest resolution are illustrated, and other branches are omitted for clarity. ``Kpts'' means keypoints.}
	\label{fig:architecture}
\end{figure*}

\subsection{Searching Process}
\label{sec:train_supernet}
We train and search BodyNet, FaceHead, and HandHead simultaneously. BodyNet is searched on four detailed architectural configuration spaces, \ie depth, width, group number, and input resolution. FaceHead/HandHead are searched on both four architectural configurations and two connection search spaces, \ie feature stage and RoI expansion. Note that the configurations of FaceHead and HandHead are not enforced to be the same, because they may prefer different architecture design.

As detailed below, the whole searching process can be divided into two stages, super-network training and sub-network searching.

\subsubsection{Super-network Training}
We train the weight-shared super-network following~\cite{cai2020once,yu2020bignas}. Sandwich rule~\cite{yu2019universally,yu2020bignas} and in-place distillation~\cite{yu2019universally,yu2020bignas} are applied, which are introduced below.

\textbf{Sandwich rule} samples the biggest sub-network, the smallest sub-network, and $N$ randomly sampled sub-networks ($N$ = 2 in our experiments) for each mini-batch. Each sampled sub-network predicts the targets and is supervised by the ground truth. As the sub-networks share weights with the super-network, the performance of all the architectures is improved in the process. Sandwich rule aims at simultaneously upgrading the upper bound and lower bound of the super-network, since the biggest and smallest sub-networks are trained in each mini-batch. 

\textbf{In-place distillation} transfers knowledge from the biggest sub-network to other sub-networks. Specifically, the heatmaps predicted by the biggest sub-network serve as the labels to supervise the smallest and randomly sampled sub-networks. In-place distillation reduces the performance gap between the biggest sub-network and the others. Note that the output heatmaps of different sub-networks might have different resolutions. To handle this mismatch in resolution, the heatmap labels from the teacher are resized according to the resolution of the sampled sub-network. 

\subsubsection{Sub-network Searching}
After training the super-network, we search the best sub-network under the given computational complexity constraint.
Specifically, we sample the architecture and connection configurations from the super-network. 
If the sampled sub-network fulfills the overall computational complexity constraint, we evaluate the performance on the validation set. The sub-network with the highest \emph{whole-body} average precision are selected as the discovered model. Instead of setting constraints on BodyNet/FaceHead/HandHead separately, \textbf{automatic computation allocation} is conducted. The search process sets the computational complexity constraint on the whole sub-network and automatically finds out the best allocation of computational costs among different sub-modules to achieve global optimum.

Fig.~\ref{fig:architecture} shows the architecture of ZoomNAS obtained by searching. Distinct architectures with different computational costs are discovered for BodyNet/FaceHead/ HandHead. We notice that ZoomNAS allocates 70\% Flops for BodyNet. This is because BodyNet estimates body/foot keypoints and provides input features for FaceHead/HandHead, which is critical for whole-body performance. We find that features at the earlier stage are selected as the input of HandHead. Since the hand areas are always smaller than face areas, low-level features containing more detailed appearance information are beneficial to hand keypoint localization. Also, HandHead chooses the larger RoI expansion than FaceHead. This is because hand keypoint localization is challenging due to occlusion, hand-hand interaction, and hand-object interaction, and more contextual information is needed. 

After the optimal sub-network is obtained, we retrain it for better performance. The training process is the same as ZoomNet and the joint training strategy is applied as introduced in Sec.~\ref{sec:train_zoomnet}.

\section{COCO-WholeBody Dataset}

COCO-WholeBody is the first large-scale 2D whole-body pose estimation dataset, which provides dense landmark annotations including 133 keypoints on the human body, feet, face, and hands. This proposed benchmark is built upon the multi-person body pose estimation dataset, COCO~\cite{lin2014microsoft}. In COCO-WholeBody, more than 200K instances are labeled and two subsets, WholeBody-Face (WBF) and WholeBody-Hand (WBH), are constructed for facial landmark localization and hand keypoint estimation.

\begin{figure}[tb]
	\centering
	\includegraphics[width=0.35\textwidth]{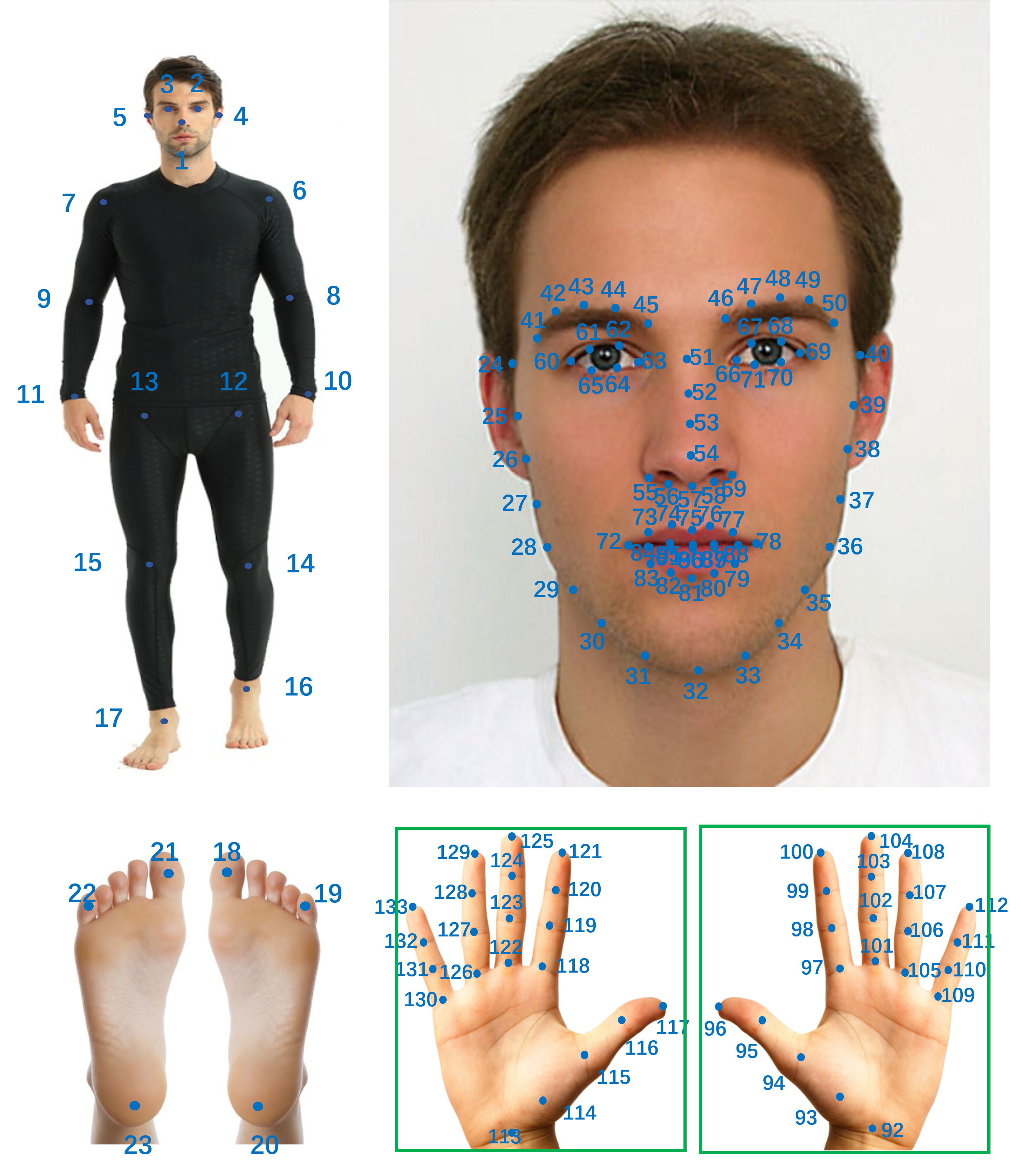}
	\caption{Definitions of 133 keypoints on COCO-WholeBody. }
	\label{fig:anno_kpt}
\end{figure}

\subsection{Data Annotation}

\begin{figure*}[tb]
	\centering
	\includegraphics[width=0.9\textwidth]{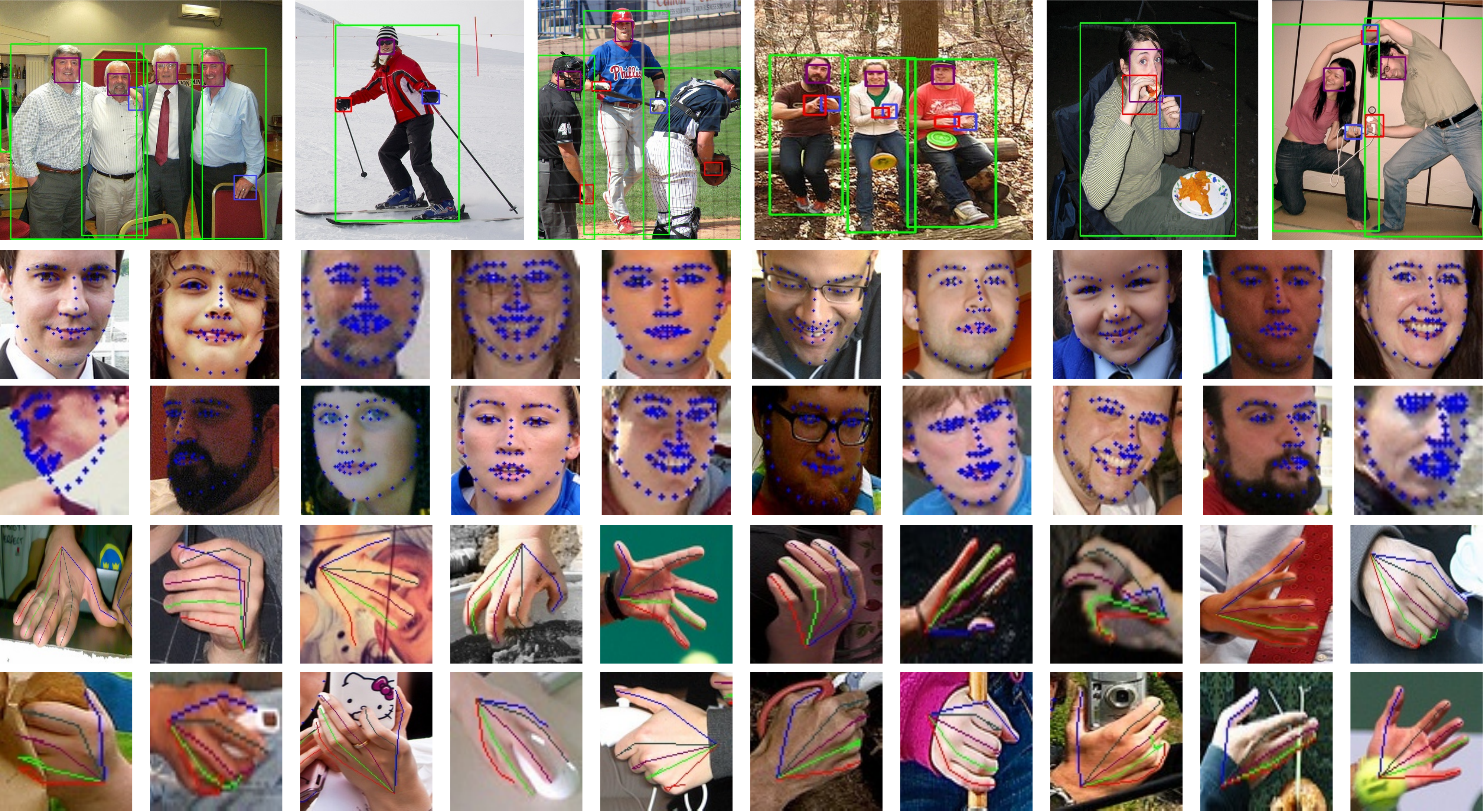}
	\caption{Annotation examples. Line \#1: Different colors are used to distinguish different types of bounding boxes, \ie body (green), face (purple), left hand (blue) and right hand (red). Line \#2 and \#3: Face keypoints. Line \#4 and \#5: Hand keypoints.}
	\label{fig:anno_hand_face_example}
\end{figure*}

For each image on the train/val set of COCO~\cite{lin2014microsoft}, we annotate the keypoints of feet, face, and hands. Combined with the original human body annotations provided by COCO, COCO-WholeBody contains dense landmark annotations on the entire human body. For each person, we provide 4 types of bounding boxes (person box, face box, left-hand box, and right-hand box) and 133 keypoints including 17 for the body, 6 for feet, 68 for face, and 42 for hands. The definitions of these  133 keypoints are illustrated in Fig.~\ref{fig:anno_kpt}.

Labeling whole-body landmarks for in-the-wild images is expensive and tedious, especially on massive and dense hand and facial keypoints.
As a rough estimation, the labeling cost of a professional annotator is up to: 10 minutes per face and 1.5 minutes per hand. The labeling cost for the face and hand bounding boxes is less than keypoint annotation, which needs 10 seconds for each bounding box on average. 
To reduce the labeling cost, we apply a semi-automatic methodology for the annotations of face and hand keypoints. Firstly, face and hand bounding boxes are annotated manually. Secondly, we use a facial landmark detection model and a hand keypoint prediction model, which are trained on large-scale face datasets and hand datasets respectively, to pre-annotate the face and hand keypoints. Thirdly, the validity of face and hand keypoints is labeled by annotators. The applied semi-automatic methodology significantly speeds up the annotation process, with $89\%$ time reduction.

To ensure the annotation quality, we follow well-defined standards to conduct quality inspections at every labeling stage. For example, the face and hand boxes are defined as the minimal bounding rectangle of the keypoints. As there are many blurry and occluded samples in our challenging dataset, the validity of these boxes is also manually annotated. They are \emph{valid} only if the face and hand images are clear enough for keypoint labeling. For the annotation of face and hand keypoints, a face model for 68 facial keypoints and a hand model for 21 hand keypoints are applied respectively on those \emph{valid} boxes. Human annotators are asked to annotate the validity of keypoints according to the quality of the pseudo-keypoints produced by the models. Some annotation examples are shown in Fig.~\ref{fig:anno_hand_face_example}.

Foot keypoints are manually labeled directly, since the labeling cost is relatively small. The big toe, the small toe, and the heel are labeled for each foot. We perform double inspections and manual corrections by asking an independent group of annotators to check and revise the same image to guarantee the data quality.

\subsection{COCO-WholeBody V1.0}
\label{sec:data_v1.0}

\begin{table}[h]
    \centering
    \caption{Compare the numbers of \emph{valid} face/hand samples in the training set of COCO-WholeBody V0.5 and V1.0.}
    \begin{tabular}{c|c|c|c}
    \hline
    Data Version & Face & Left-hand & Right-hand \\
    \hline
    V0.5~\cite{jin2020whole} & 53,657 & 19,145 & 20,133 \\
    V1.0 & 53,877 & 37,954 & 40,845 \\ 
    \hline
    \end{tabular}
    \label{tab:valid_num}
\end{table}

We introduce the COCO-WholeBody dataset generated in our preliminary work~\cite{jin2020whole} as V0.5, which labels the validity of keypoints at the whole face/hand level and still contains many \emph{invalid} face/hand. In this work, we further improve the annotations of COCO-WholeBody from V0.5 to V1.0, and make it publicly available\footnote{https://github.com/jin-s13/COCO-WholeBody}. For these \emph{invalid} face/hand keypoints that are discarded during quality inspections on the original COCO-WholeBody V0.5 training set, we hire professional annotators to manually re-annotate the keypoints if possible. As a result, the number of \emph{valid} samples of hands increases by 100\% as shown in Table~\ref{tab:valid_num}. For these originally \emph{valid} face/hand on COCO-WholeBody V0.5, strict quality inspections are performed at the joint level and manual corrections are conducted for incorrect keypoints, which upgrades the annotation quality in the challenging scenarios. As incorrect/missing labels of face/hand have already been corrected on COCO-WholeBody V0.5 validation set, it remains the same for COCO-WholeBody V1.0. In this way, both quantity and quality of the annotations are improved, which promotes the performance of the data-driven deep learning approaches as validated in Sec.~\ref{sec:ablation_study}.

\begin{figure}[tb]
	\centering
	\includegraphics[width=0.48\textwidth]{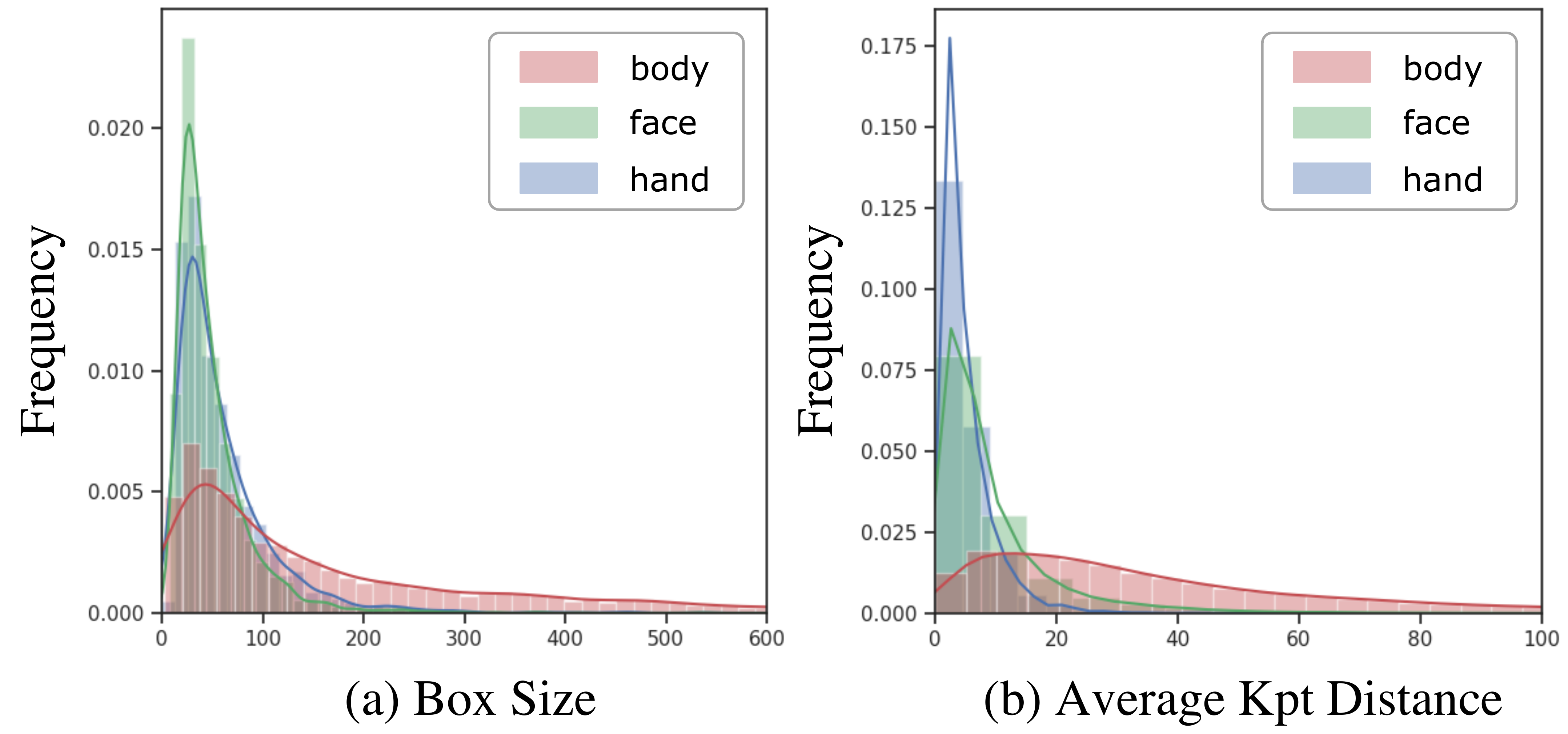}
	\caption{(a) The box size of body/face/hand (b) The average keypoint distance of body/face/hand. We notice the large scale variance of different parts of the same person.}
	\label{fig:box_kpt_distance}
\end{figure}

\begin{figure}[tb]
	\centering
	\includegraphics[width=0.48\textwidth]{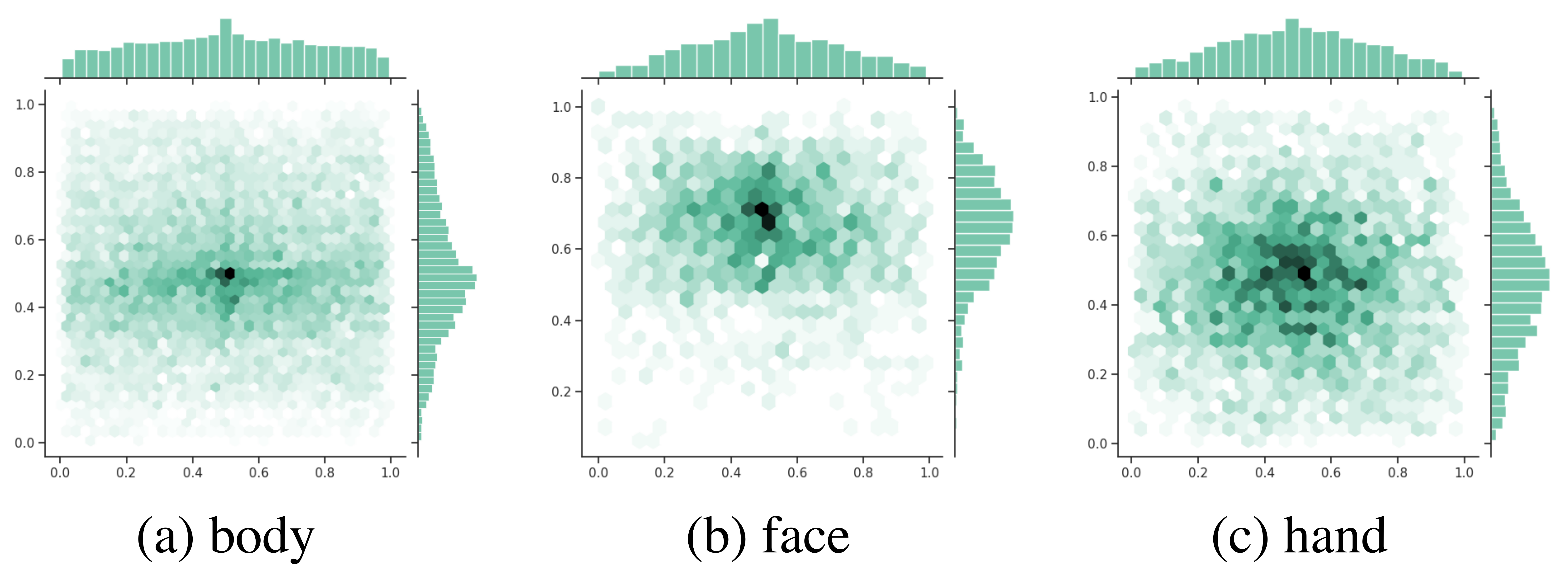}
	\caption{Distribution of the center location over the whole image (a) body, (b) face, and (c) hand.}
	\label{fig:center_dist}
\end{figure}

We analyze the statistics of COCO-WholeBody V1.0.
In Fig.~\ref{fig:box_kpt_distance}, we plot the distributions of two aspects, (a) the box size and (b) the average keypoint distance of different body parts. The box size is calculated as the diagonal length (in pixel) of body, face, and hand boxes respectively. The keypoint distance is calculated as the average distance between the tree-structured keypoint pairs. We notice that hand and face have obviously smaller scales than body. Fig.~\ref{fig:center_dist} illustrates the distributions of the center location of human body, face, and hand with respect to the image. The center location is normalized by the image width and height. We can see that the various scale and position distributions make it challenging to localize keypoints of different body parts simultaneously. 

Overall, COCO-WholeBody V1.0 is a large-scale dataset with whole-body annotations, which will promote research on whole-body pose estimation, as well as contribute to other related areas such as face and hand detection, face alignment, and hand pose estimation.

\begin{figure}[tb]
	\centering
	\includegraphics[width=0.4\textwidth]{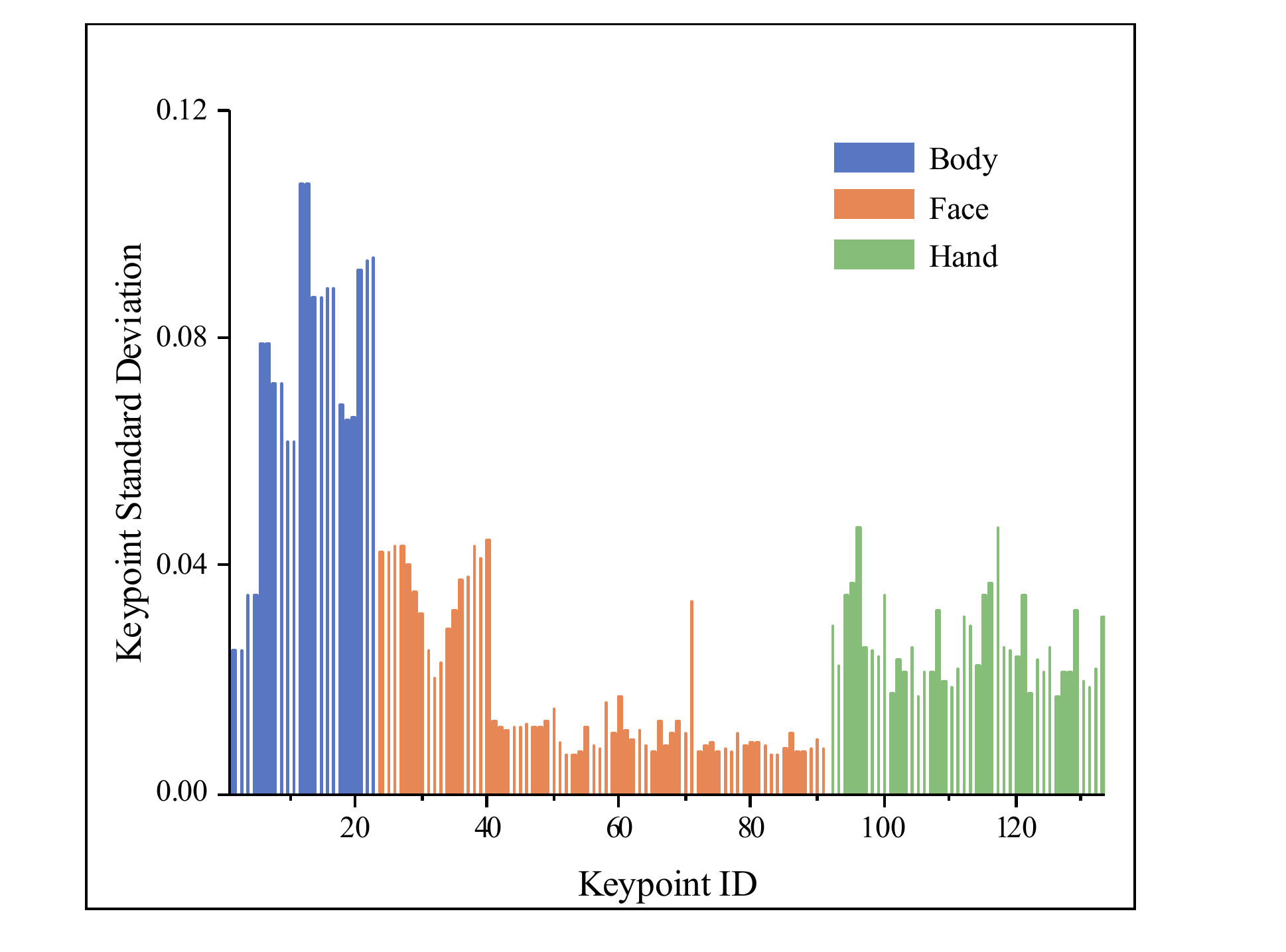}
	\caption{The normalized standard deviation of manual annotation for each keypoint. Body keypoints have the larger manual annotation variance than face and hand keypoints.}
	\label{fig:standard_deviation}
\end{figure}

\textbf{Evaluation Metric.}
Following~\cite{lin2014microsoft}, we employ mean Average Precision (mAP) and Average Recall (mAR) as the evaluation metric, where Object Keypoint Similarity (OKS) is used to measure the similarity between the prediction and the ground truth poses. OKS is defined as follows,

\begin{equation}
    \operatorname{OKS} = \frac{\sum_{i}\exp(-d_i^2/2s^2k_i^2)\delta(v_i > 0)}{\sum_i \delta(v_i > 0)},
\label{eq:oks}
\end{equation}

\noindent where $d_i$ is the Euclidean distance between each ground truth and the detected keypoint. $v_i$ is the visibility flag. $s$ is the scale of the person. $k_i$ is a per-keypoint constant to control falloff. The constant is a normalized factor that indicates the fault tolerance for each type of keypoint. For body keypoints, we directly use $k_i$ provided by COCO~\cite{lin2014microsoft}. For other keypoints, the normalized factor is calculated according to the standard deviation of the human annotations. The same batch of 500 images containing face/hand/foot keypoints are labeled by three different annotators and the standard deviation is calculated for each keypoint. The normalized standard deviation is shown in Fig.~\ref{fig:standard_deviation}.

For the task of whole-body pose estimation, we report the metric for whole-body as well as body/foot/face/hand. The performance of whole-body pose estimation measures the accuracy of all 133 keypoints at the instance level, while that of body/foot/face/hand is evaluated for the corresponding human parts individually.

\subsection{WholeBody-Face (WBF) Dataset}

\begin{figure}[tb]
	\centering
	\includegraphics[width=0.48\textwidth]{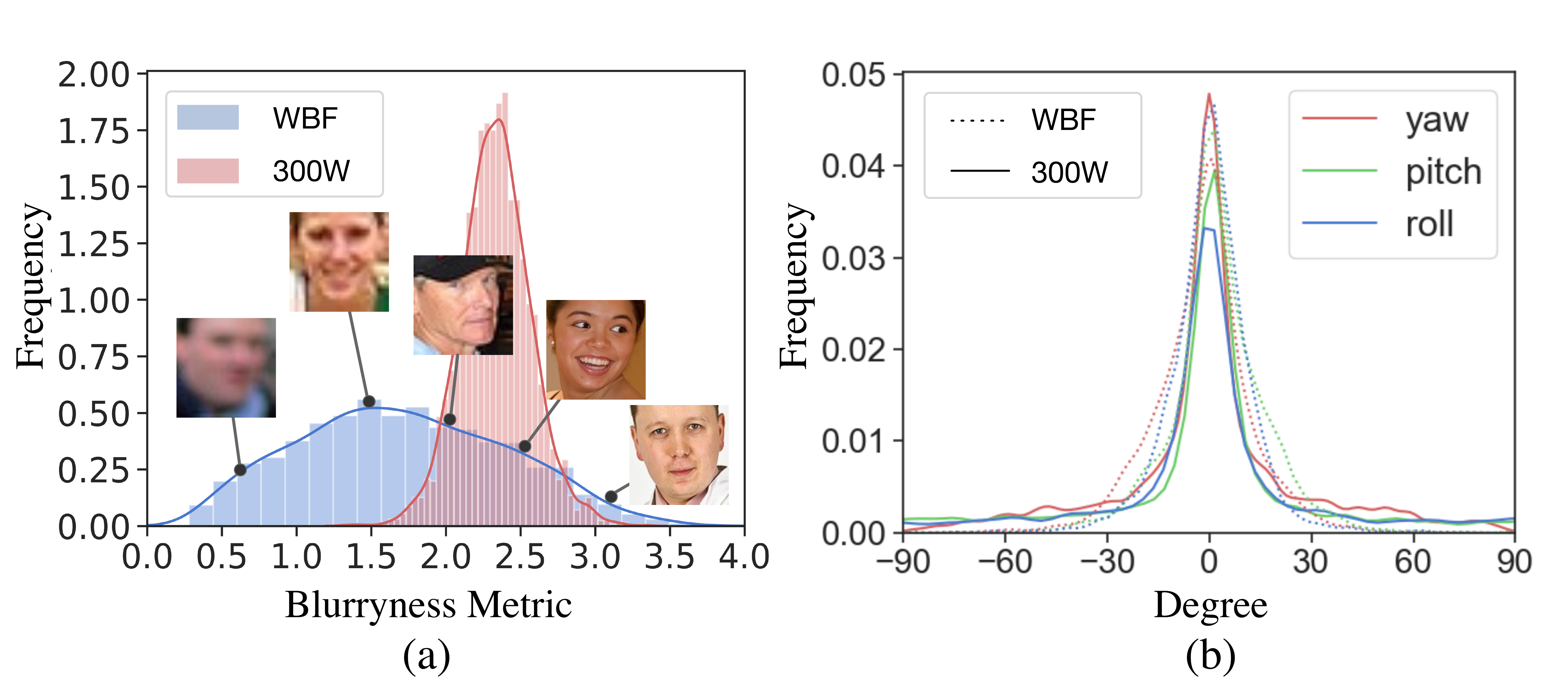}
	\caption{WBF is challenging as it contains (a) many blurry face images and (b) large head pose variations.}
	\label{fig:face_compare}
\end{figure}

We present a sub-dataset, WholeBody-Face (WBF), for facial landmark detection. WBF is created by extracting cropped faces from COCO-WholeBody V1.0. More than 50K training images and 2K validation images are provided. Following 300W~\cite{sagonas2013300}, 68 keypoints are annotated for each face. 

We compare WBF and 300W~\cite{sagonas2013300} in Fig.~\ref{fig:face_compare} and find that our proposed WBF is more challenging than 300W. In Fig.~\ref{fig:face_compare}(a), we apply Laplacian method~\cite{Pech2000Diatom} to measure the \emph{blurriness} of face images. The images of 300W are clear and the calculated blurriness is mainly in an interval between $1$ and $3$. In comparison, WBF has a larger variance of blurriness and contains more challenging images (blurriness $<1$). We also estimate the head pose by fitting the annotated 2D facial landmarks to a generic 3D face model. The distribution of the head pose (yaw, pitch, roll) is illustrated in Fig.~\ref{fig:face_compare}(b). We notice that WBF has a large variance of head poses similar to 300W dataset.

\textbf{Evaluation Metric.}
We use the normalized mean error (NME) to evaluate the accuracy of predicted face keypoints,
\begin{equation}
    \mathrm{NME} = \frac{1}{N} \sum_{i=1}^{N} \frac{\lVert p_i - \hat{p}_i \rVert_2}{d},
\end{equation}
where $p_i$ is the predicted keypoint location and $\hat{p}_i$ is its ground truth location. $N$ is the number of face keypoints. $d$ denotes the normalize factor. We use the inter-ocular distance as the normalize factor, which is measured as the Euclidean distance between the outer corners of the eyes. 

\subsection{WholeBody-Hand (WBH) Dataset}
We build a hand keypoint dataset, WholeBody-Hand (WBH), by extracting cropped hands from COCO-WholeBody V1.0. WBH has 21-keypoint annotations for each hand. The training set contains around 80K images, with 4K images for the validation set. Unlike most previous hand datasets that are collected in constrained environments, our WBH is the largest in-the-wild dataset for 2D hand keypoint localization.

\begin{figure}[tb]
	\centering
	\includegraphics[width=0.45\textwidth]{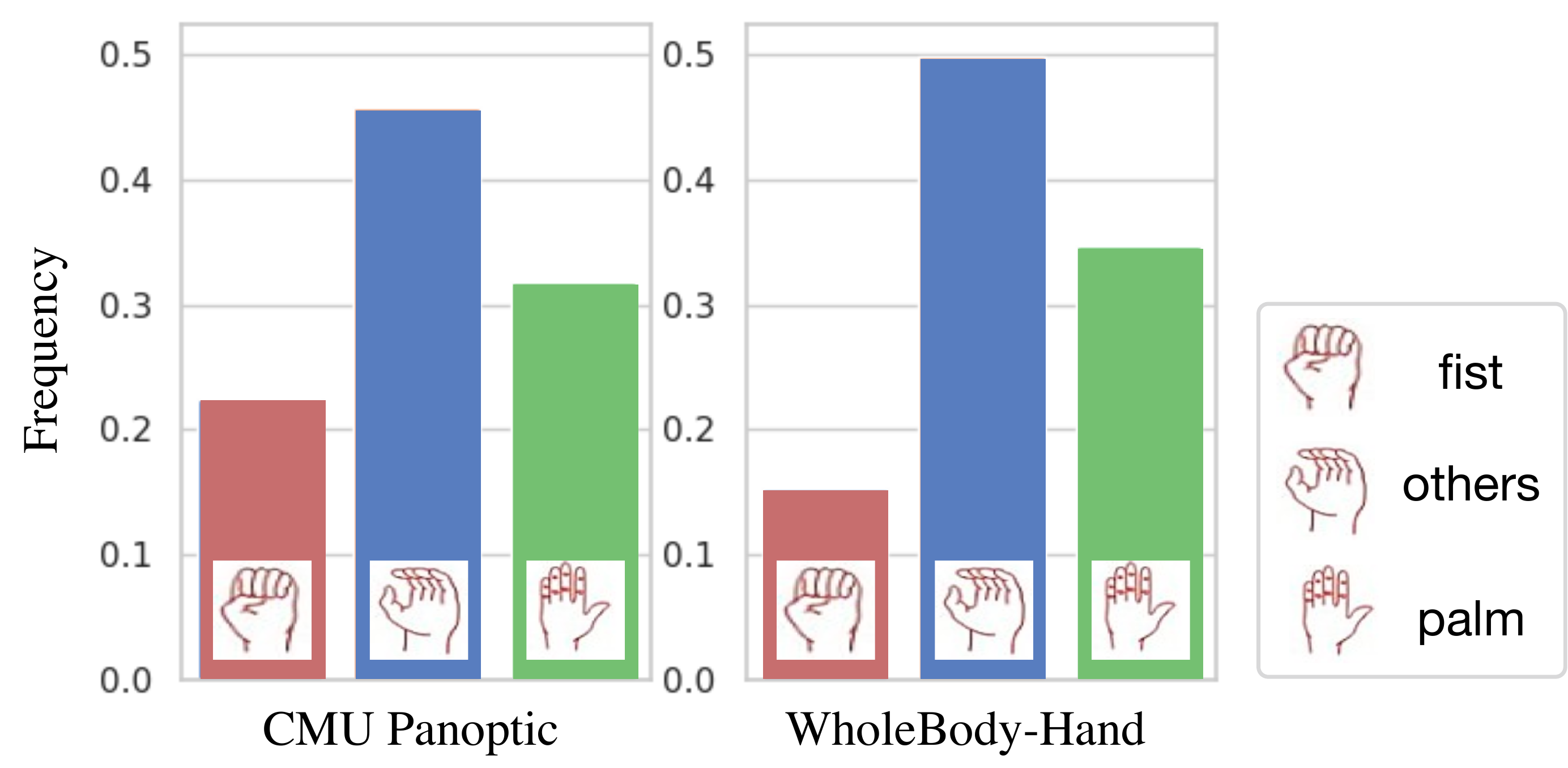}
	\caption{WBH is challenging as it contains more complex hand poses except the standard hand gestures.}
	\label{fig:hand_pose}
\end{figure}

We analyze the hand gesture variance of our proposed WBH dataset in Fig.~\ref{fig:hand_pose}. The 2D hand poses are normalized by rotating and scaling, and then are clustered into three categories: \emph{fist}, \emph{palm} and \emph{others}. Compared with Panoptic~\cite{gomez2017large}, WholeBody-Hand is more challenging as it contains a larger proportion of hand images grasping or holding objects rather than the standard hand gestures. 

\textbf{Evaluation Metric.}
We use the percentage of correct keypoint (PCK), area under the curve (AUC), and average endpoint error (EPE) for evaluation.

\textit{PCK} is a measurement between the predicted and ground truth keypoint locations within a normalized distance threshold $\sigma$ ($\sigma = 0.2$ in our experiments),
\begin{equation}
    \mathrm{PCK} = \frac{1}{N} \sum_{i=1}^{N} \mathbf{1} \left( \frac{ \lVert p_i - \hat{p}_i \rVert_2}{\max(w, h)} \leq \sigma \right),
\end{equation}
where $p_i$ is the predicted keypoint location and $\hat{p}_i$ is its ground truth location. $N$ is the number of keypoints. $\mathbf{1}(\cdot)$ is the indicator function. $w$ and $h$ are the width and height of the ground truth bounding box.

\textit{AUC} is the area under the percentage of correct keypoint curve. The predicted keypoint is considered correct, when it falls within the certain absolute distance threshold compared with the ground truth keypoint. AUC calculates the integral of the correct percentage on the interval of thresholds from 0 to 30 pixels.

\textit{EPE} is the average endpoint error in pixels, \ie the average Euclidean distance between the predicted keypoints and the ground truth keypoints.

\section{Experiments}

\begin{table*}[t]
	\caption{Whole-body pose estimation results on COCO-WholeBody V1.0 dataset. Besides the existing whole-body pose estimation methods, we build upon the top-down and bottom up body pose estimation methods and report the whole-body performance.  Our proposed ZoomNet and ZoomNAS outperforms these methods. `*' indicates multi-scale testing.}
	\label{tab:compare}
	\begin{center}
		\begin{tabular}{c|c|c|cc|cc|cc|cc|aa}
			\hline
			& \multirow{2}{*}{Method} & \multirow{2}{*}{GFLOPs} & \multicolumn{2}{c|}{body}  & \multicolumn{2}{c|}{foot}  & \multicolumn{2}{c|}{face}  & \multicolumn{2}{c|}{hand} & \multicolumn{2}{c}{\cellcolor{Gray}whole-body} \\
			\cline{4-13}
			& ~ & ~ &  AP     & AR     & AP   & AR     &  AP  & AR     & AP    & AR   &  AP     & AR  \\
			\hline
            \multirow{2}{*}{Whole-body} & SN$^{*}$~\cite{hidalgo2019single} & 272.30 & 42.7 & 58.3 & 9.9 & 36.9 & 64.9 & 69.7 & 40.8 & 58.0 & 32.7 & 45.6 \\ 
            ~ & OpenPose~\cite{cao2018openpose} & 451.09 & 56.3 & 61.2 & 53.2 & 64.5 & 76.5 & 84.0 & 38.6 & 43.3 & 44.2 & 52.3 \\ \hline
			\multirow{2}{*}{Bottom-up} & PAF$^{*}$~\cite{cao2017realtime} & 329.12 & 38.1 & 52.6 & 5.3 & 27.8 & 65.5 & 70.1 & 35.9 & 52.8 & 29.5 & 40.5 \\ 
            ~ & AE~\cite{newell2017associative} & 212.36 & 58.0 & 66.1 & 57.7 & 72.5 & 58.8 & 65.4 & 48.1 & 57.4 & 44.0 & 54.5 \\ \hline
            \multirow{4}{*}{Top-down} & DeepPose~\cite{toshev2014deeppose} & 17.30 & 44.4 & 56.8 & 36.8 & 53.7 & 49.3 & 66.3 & 23.5 & 41.0 & 33.5 & 48.4 \\
            ~ & SimpleBaseline~\cite{xiao2018simple} & 20.42 & 66.6 & 74.7 & 63.5 & 76.3 & 73.2 & 81.2 & 53.7 & 64.7 & 57.3 & 67.1 \\
			~ & HRNet~\cite{sun2019deep} & 16.04 & 70.1 & 77.3 & 58.6 & 69.2 & 72.7 & 78.3 & 51.6 & 60.4 & 58.6 & 67.4 \\ 
			~ & PVT~\cite{wang2021pyramid} & 19.65 & 67.3 & 76.1 & 66.0 & 79.4 & 74.5 & 82.2 & 54.5 & 65.4 & 58.9 & 68.9 \\ \hline
			\multirow{2}{*}{Ours} & ZoomNet & 28.49 & 74.5 & 81.0 & 60.9 & 70.8 & 88.0 & 92.4 & 57.9 & 73.4 & 63.0 & 74.2 \\
			~ & ZoomNAS & 18.02 & 74.0 & 80.7 & 61.7 & 71.8 & 88.9 & 93.0 & 62.5 & 74.0 & 65.4 & 74.4 \\
			\hline
		\end{tabular}
	\end{center}
\end{table*}

\subsection{COCO-WholeBody V1.0 Dataset}

We present the benchmarking results of whole-body pose estimation including our proposed methods on COCO-WholeBody V1.0 dataset. There are two existing approaches for 2D whole-body pose estimation, \ie Single-Network (SN)~\cite{hidalgo2019single} and OpenPose~\cite{cao2018openpose}. To extensively evaluate the performance of existing methods on the proposed dataset, we also build strong baseline methods based on the body pose estimation approaches, including both bottom-up methods (\ie Part Affinity Fields (PAF)~\cite{cao2017realtime} and Associate Embedding (AE)~\cite{newell2017associative}) and top-down methods (\ie DeepPose~\cite{toshev2014deeppose}, SimpleBaseline~\cite{xiao2018simple}, HRNet~\cite{sun2019deep}, and Pyramid Vision Transformer (PVT)~\cite{wang2021pyramid}), and adapt them to the whole-body pose estimation task. We evaluate the proposed ZoomNet and ZoomNAS, and demonstrate their superiority over existing methods.

\subsubsection{Baseline Methods} 
The implementation details of the baseline methods are listed as follows:

\textbf{Whole-body Pose Estimation Methods.}
\textit{SN}~\cite{hidalgo2019single} is a recently proposed whole-body pose estimation approach. We retrain the whole-body keypoint estimator on our proposed COCO-WholeBody V1.0 dataset. All the keypoints are supervised simultaneously since COCO-WholeBody V1.0 provides whole-body annotations. The number of heatmaps is 133, and the number of part-affinity fields (PAFs) is 135 as we design a tree structure except for 3 loops.
We use 5 stages for PAFs and 1 stage for heatmaps. \textit{OpenPose}~\cite{cao2018openpose} is a multi-network whole-body pose estimation system, which consists of a body keypoint model, a facial landmark detector, and a hand pose estimator. We reimplement the approach by training these models separately on COCO-WholeBody V1.0 dataset.

\textbf{Bottom-up Body Pose Estimation Methods.}
\textit{PAF}~\cite{cao2017realtime} adopts PAFs for pose grouping and applies CPM~\cite{wei2016convolutional} as the backbone. We use the same training setting as SN introduced above for whole-body pose estimation. \textit{AE}~\cite{newell2017associative} learns to group keypoints by associative embedding, which is flexible in terms of various numbers of predicted keypoints. We use the 4-stacked hourglass backbone and follow the same training settings as~\cite{newell2017associative} in our experiments.

\textbf{Top-down Body Pose Estimation Methods.}
\textit{DeepPose}~\cite{toshev2014deeppose} is a regression-based body keypoint estimation method, with ResNet-101~\cite{he2016deep} as the backbone. \textit{SimpleBaseline}~\cite{xiao2018simple} is a recently proposed top-down method for human body pose estimation, which adopts ResNet-50~\cite{he2016deep} as the backbone and uses a few deconvolutional layers as the head. \textit{HRNet}~\cite{sun2019deep} is the state-of-the-art approach for human pose estimation. We choose HRNet-W32 as the backbone, which is the same as ZoomNet. \textit{
PVT}~\cite{wang2021pyramid} is a transformer-based backbone designed for dense prediction vision tasks. Deconvolutional layers are applied as the keypoint head for whole-body pose estimation. Following the top-down paradigm, these methods first detect all the person candidates and then estimate the whole-body keypoints for each person. We retrain the models to fit in the whole-body pose estimation task by increasing the number of keypoints to 133. For human detection, we use the same detected human bounding boxes provided by SimpleBaseline~\cite{xiao2018simple}.

\subsubsection{Implementation Details} 
We train and test our proposed ZoomNet and ZoomNAS on COCO-WholeBody V1.0 dataset. We follow the same data augmentation as MMPose~\cite{mmpose2020} with random scaling ([-50\%, 50\%]), random rotation ([$-40^{\circ}$, $40^{\circ}$]), and random flipping. The backbone networks are pre-trained on ImageNet dataset. 2D gaussian heatmaps with $\sigma=3$ are used to encode the keypoint locations. 

For ZoomNet, we resize the cropped person images to $384 \times 288$ as the input of BodyNet. The input features of FaceHead/HandHead are cropped from the first stage of BodyNet according to the predicted face/hand boxes without enlargement. The cropped features are resized to $64 \times 64$ and are processed by FaceHead/HandHead for face/hand keypoint localization. BodyNet and FaceHead/HandHead are jointly trained on 16 GPUs with a batch size of 16 in each GPU for 210 epochs. We use Adam optimizer~\cite{kingma2014adam} with the base learning rate of 1e-3, and decay the learning rate to 1e-4 and 1e-5 at the 170th and 200th epochs respectively. 

For ZoomNAS, we train the super-network for 210 epochs with the same learning schedule. 
Then we sample 500 sub-networks under the overall Flops constraints and find out the neural architecture with the highest whole-body AP. The obtained architecture is retrained with the same training setting as ZoomNet.

\subsubsection{Benchmarking Comparison}
Table~\ref{tab:compare} compares our proposed ZoomNet and ZoomNAS with the baseline methods on COCO-WholeBody V1.0 dataset. We also report the computational complexity (GFLOPs) of all models. For bottom-up models, the complexity is computed under the single-scale testing configuration. And for top-down methods, the complexity excludes that of the human detection model.

ZoomNet and ZoomNAS significantly outperform the existing whole-body pose estimation approaches, \ie SN~\cite{hidalgo2019single} and OpenPose~\cite{cao2018openpose}. SN follows a one-stage paradigm, which predicts all 133 keypoints simultaneously. We find that localizing body/foot/face/hand keypoints at once is challenging for network training. The applied bottom-up pipeline ignores the scale variation of different persons and different body parts, leading to performance degradation. OpenPose applies multiple heavy models to estimate keypoints of different body parts individually, which is computationally inefficient (137.52 GFLOPs for BodyNet, 106.77 GFLOPs for FaceNet, and $103.40 \times 2=206.80$ GFLOPs for HandNet). As the body model and the face/hand models are trained independently, OpenPose does not utilize the relationship between human parts. In addition, the face/hand boxes are roughly estimated by hand-crafted rules from the estimated body keypoints, which hinders the face/hand pose estimation accuracy. In comparison, our proposed ZoomNAS conducts body/foot keypoint localization and zooms in to focus on the face/hand areas in a single network. ZoomNAS captures the relationship between human parts and jointly optimizes the whole-body keypoints. It significantly outperforms SN by 30.3\% mAP and OpenPose by 18.8\% mAP.

ZoomNet shows superiority over other baseline methods that build upon the multi-person pose estimation approaches. The popular bottom-up methods, PAF~\cite{cao2017realtime} and AE~\cite{newell2017associative}, suffer from the same problems as SN discussed above. 
Top-down methods regress the coordinates (DeepPose~\cite{toshev2014deeppose}) or predict the heatmaps (SimpleBaseline~\cite{xiao2018simple}, HRNet~\cite{sun2019deep}, and PVT~\cite{wang2021pyramid}) of whole-body keypoints all at once. HRNet~\cite{sun2019deep} can be viewed as the one-stage alternative of ZoomNet, since they share the same network backbone (HRNet-W32). ZoomNet significantly outperforms HRNet by 4.4\% mAP and 6.8\% mAR, demonstrating the effectiveness of utilizing the hierarchical structure of the full human body.

ZoomNAS further promotes both efficiency and accuracy over ZoomNet. Our proposed ZoomNAS jointly searches the neural architectures of BodyNet/FaceHead/ HandHead and the connections between these sub-modules. The computational complexity is automatically allocated to BodyNet/FaceHead/HandHead for the global optimum. Compared with ZoomNet, ZoomNAS achieves higher accuracy (65.4\% vs 63.0\% mAP) with 37\% computational complexity reduction (18.02 vs 28.49 GFLOPs).

\begin{figure*}[h]
	\centering
	\includegraphics[width=0.9\linewidth]{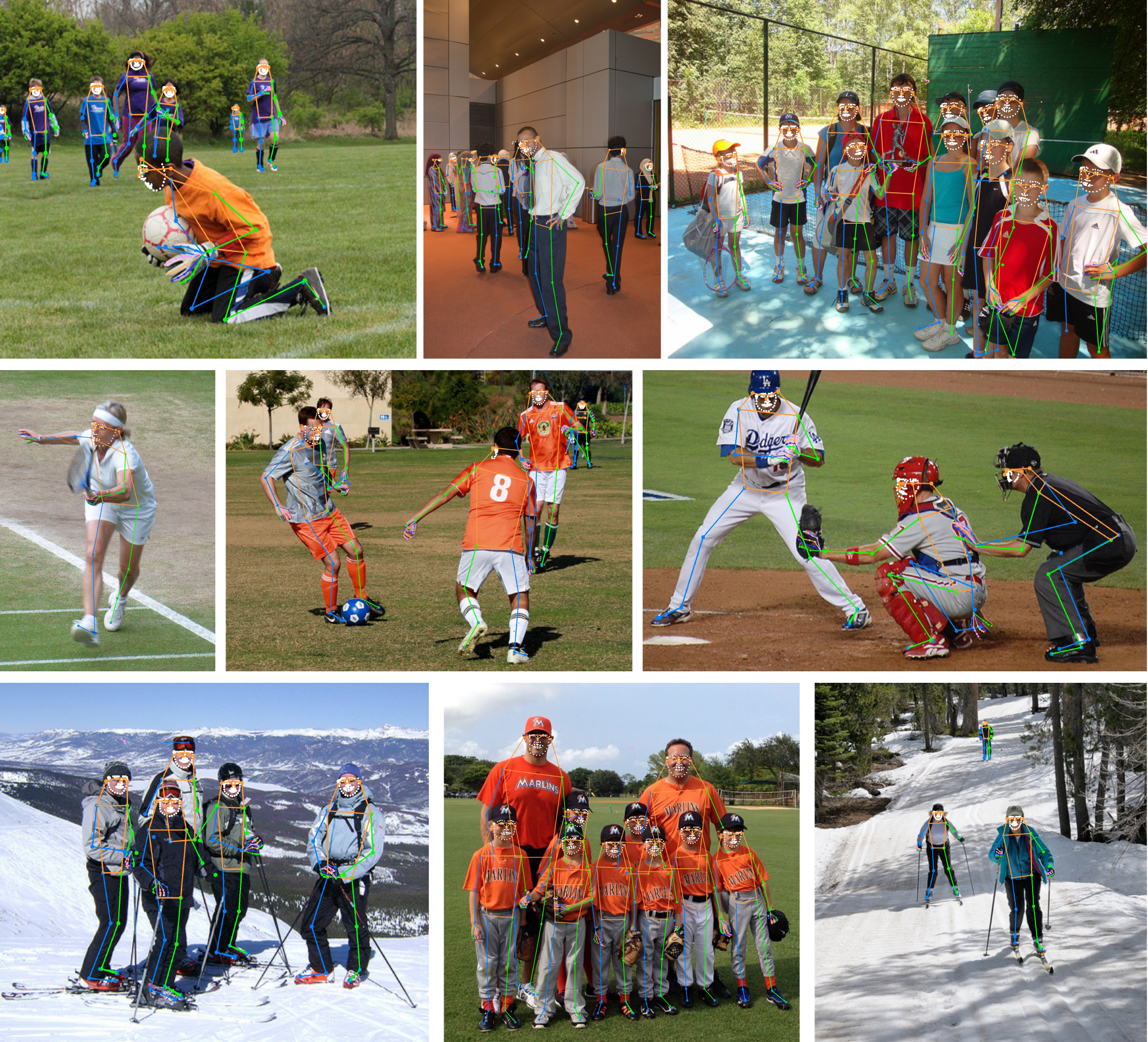}
	\caption{Qualitative evaluation results of ZoomNAS in handling challenges including scale variation, close proximity, occlusion, and pose diversity.}
	\label{fig:visualization}
\end{figure*}

\subsubsection{Qualitative Analysis}
Figure~\ref{fig:visualization} demonstrates the qualitative results of ZoomNAS. 
We show that ZoomNAS is robust to scale variation and captures the small-scale persons and human parts. It is also capable of handling challenges including close proximity, occlusion, and pose variation.

\subsection{WholeBody-Face (WBF) Dataset}

\subsubsection{Benchmarking on WBF Dataset}
\label{sec:benchmark_face}

Table~\ref{tab:benchmark_coco_wholebody_face} presents the benchmarking results on our proposed facial landmark dataset, WholeBody-Face (WBF).

\begin{table}[h]
    \centering
    \caption{Benchmarking results on WBF dataset. NME is adopted for evaluation. $\downarrow$ means lower is better.}
    \begin{tabular}{c|c}
\hline
Method & NME $\downarrow$ \\
\hline
DeepPose-ResNet50~\cite{toshev2014deeppose} & 6.45 \\
MobileNetV2~\cite{sandler2018mobilenetv2} & 6.12 \\ 
Hourglass52~\cite{law2018cornernet} & 5.86 \\
SimpleBaseline-ResNet50~\cite{xiao2018simple} & 5.66 \\
SCNet50~\cite{liu2020improving} & 5.65 \\
HRNetV2-W18~\cite{wang2020deep} & 5.68 \\ 
HRNetV2-W18~\cite{wang2020deep} + DARK~\cite{zhang2020distribution} & \textbf{5.12} \\ 
\hline
    \end{tabular}
    \label{tab:benchmark_coco_wholebody_face}
\end{table}

\textbf{Baseline Methods.}
\textit{DeepPose}~\cite{toshev2014deeppose} is a regression-based keypoint estimation method. ResNet-50~\cite{he2016deep} is applied as its backbone in our implementation.
\textit{MobileNetV2}~\cite{sandler2018mobilenetv2} is a popular mobile architecture. We use it as the backbone and replace its classification head with several deconvolutional layers to output heatmaps. \textit{Hourglass52}~\cite{law2018cornernet} is a modified version of stacked hourglass network~\cite{newell2016stacked}, which changes the number of feature channels and the down-sampling method, and modifies the feature fusion module. Please refer to~\cite{law2018cornernet} for more details about the network architecture. 
\textit{SimpleBaseline}~\cite{xiao2018simple} pose estimator adds deconvolutional layers to the backbone network. We use ResNet-50~\cite{he2016deep} as its backbone. 
\textit{SCNet}~\cite{liu2020improving} is a recently proposed backbone network, which applies the self-calibrated convolution to capture long-range spatial and inter-channel dependencies.
\textit{HRNetV2}~\cite{wang2020deep} is the state-of-the-art model for 2D facial landmark detection. It improves upon HRNet~\cite{sun2019deep} by adopting multi-resolution feature fusion in the representation head.
\textit{DARK}~\cite{zhang2020distribution} is a model-agnostic plug-in technique for unbiased coordinate encoding/decoding.

\textbf{Implementation.} Following the common setting~\cite{wang2020deep}, we use the cropped face images and resize them to $256 \times 256$. For heatmap-based methods, the output heatmap size is $64 \times 64$ and the sigma of the target Gaussian distribution is $2$. 
During training, we perform random scaling ([-25\%, 25\%]), random rotation (([$-30^{\circ}$, $30^{\circ}$]), and random flipping. The models are trained with Mean-Squared Error (MSE) loss for $60$ epochs. Adam~\cite{kingma2014adam} is adopted for optimization. The base learning rate is 4e-3, and is reduced by a factor of $10$ at the $40$-th and $55$-th epochs. During testing, we adopt the flip-testing strategy and obtain the predicted keypoint locations by adding a quarter offset in the direction from the highest response to the second-highest response~\cite{chen2018cascaded}.

\begin{table}[t]
    \centering
    \caption{Cross-dataset evalutation for WBF dataset. HRNetV2-W18 is pretrained on WBF and the effect is tested on benchmark datasets. NME is adopted for evaluation.}
    \begin{tabular}{c|c|c|c|c}
\hline
 & WFLW & AFLW & COFW & 300W \\
\hline
w/o pretrain & 4.60 & 1.57 & 3.45 & 3.32  \\
pretrain on WBF & 4.01 & 1.42 & 3.36 & 3.21 \\
\hline
    \end{tabular}
    \label{tab:pretrain_WBF}
\end{table}

\begin{table*}[ht]
\centering
\caption{Cross-dataset generalization results measured by PCK$\uparrow$/AUC$\uparrow$/EPE$\downarrow$. We choose one dataset for training, and the others for testing. The row represents the training set and the column represents the testing set. \textcolor{red}{\textbf{\underline{RED}}} indicates the best performance and \textcolor{blue}{\textbf{BLUE}} indicates the second-best performance on the testing set respectively.}
\begin{tabular}{l|c|c|c|c}
\hline
\backslashbox[27mm]{Train}{Test} & Panoptic~\cite{simon2017hand} & OneHand10K~\cite{Yangang2018Mask} & FreiHand~\cite{Freihand2019} & WBH \\
\hline

Panoptic~\cite{simon2017hand} & \textcolor{red}{\textbf{\underline{99.9}}}/\textcolor{blue}{\textbf{72.7}}/\textcolor{blue}{\textbf{8.46}} & 84.1/37.0/104.13 & 81.7/59.7/13.20 & \textcolor{blue}{\textbf{65.6}}/\textcolor{blue}{\textbf{76.0}}/\textcolor{blue}{\textbf{7.18}} \\ \hline 

OneHand10K~\cite{Yangang2018Mask} & 99.4/59.4/14.54 & \textcolor{red}{\textbf{\underline{98.7}}}/\textcolor{red}{\textbf{\underline{55.7}}}/\textcolor{red}{\textbf{\underline{26.71}}} & 79.2/57.9/14.35 & 58.1/73.6/7.91 \\ \hline

FreiHand~\cite{Freihand2019} & 99.2/51.3/18.86 & 67.8/28.4/183.00 & \textcolor{red}{\textbf{\underline{99.2}}}/\textcolor{red}{\textbf{\underline{86.8}}}/\textcolor{red}{\textbf{\underline{3.25}}} & 41.1/64.0/11.54 \\ \hline 

WBH & \textcolor{blue}{\textbf{99.8}}/\textcolor{red}{\textbf{\underline{73.6}}}/\textcolor{red}{\textbf{\underline{8.07}}} & \textcolor{blue}{\textbf{97.0}}/\textcolor{blue}{\textbf{47.5}}/\textcolor{blue}{\textbf{47.47}} & \textcolor{blue}{\textbf{86.6}}/\textcolor{blue}{\textbf{65.0}}/\textcolor{blue}{\textbf{11.02}} & \textcolor{red}{\textbf{\underline{80.2}}}/\textcolor{red}{\textbf{\underline{83.5}}}/\textcolor{red}{\textbf{\underline{4.58}}} \\ \hline

\end{tabular}
\label{tab:hand_cross}
\end{table*}

\subsubsection{Cross-dataset Evaluation}
We show that WBF is complementary to other face keypoint datasets by evaluating its generalization ability in Table~\ref{tab:pretrain_WBF}.

\textbf{Benchmark Datasets.} We use four benchmark datasets.
(1) WFLW~\cite{wu2018look} dataset contains $10$k faces ($7.5$k for training and $2.5$k for testing) with $98$ annotated landmarks. 
(2) AFLW~\cite{koestinger2011annotated} dataset consists of $20$K training images with $19$ annotated landmarks. The evaluation is conducted on AFLW-Full set ($4,386$ images).
(3) COFW~\cite{Burgos2013Robust} dataset consists of $1,345$ training and $507$ testing faces with $29$ annotated landmarks.
(4) 300W~\cite{sagonas2013300} dataset contains $3,148$ faces with $68$ landmarks for training. The evaluation is performed on the full test set ($689$ faces).

We use HRNetV2-W18~\cite{wang2020deep} for facial landmark detection, which is the state-of-the-art model for face keypoint localization. Normalized mean error (NME) with inter-ocular distance as the normalization factor is adopted for evaluation. We find that WBF is complementary to other face datasets, and pretraining on WBF further improves upon the state-of-the-art performance on these datasets.

\subsection{WholeBody-Hand (WBH) Dataset}

\begin{table}[h]
    \centering
    \caption{Benchmarking results on WBH dataset. $\uparrow$ means higher is better, while $\downarrow$ means lower is better.}
    \begin{tabular}{c|c|c|c}
\hline
Method & PCK $\uparrow$ & AUC $\uparrow$ & EPE $\downarrow$ \\
\hline
DeepPose-res50~\cite{toshev2014deeppose} & 77.7 & 80.7 & 5.33 \\
MobileNetV2~\cite{sandler2018mobilenetv2} & 77.0 & 81.5 & 5.24 \\
SimpleBaseline-res50~\cite{xiao2018simple} & 79.4 & 83.0 & 4.75 \\ 
Hourglass52~\cite{law2018cornernet} & 80.0 & 83.3 & 4.61 \\
SCNet50~\cite{liu2020improving} & \textbf{80.5} & 83.5 & 4.57 \\ 
HRNetV2-W18~\cite{wang2020deep} & 80.2 & 83.4 & 4.57 \\ 
HRNetV2-W18~\cite{wang2020deep} + DARK~\cite{zhang2020distribution} & \textbf{80.5} & \textbf{83.7} & \textbf{4.48} \\ 
\hline
\end{tabular}
\label{tab:benchmark_coco_wholebody_hand}
\end{table}

\subsubsection{Benchmarking on WBH Dataset}
\label{sec:benchmark_hand}
Table~\ref{tab:benchmark_coco_wholebody_hand} demonstrates the benchmarking results of hand pose estimation on our proposed WholeBody-Hand (WBH) dataset. The benchmark experimental settings are almost the same as those of WholeBody-Face dataset in Sec.~\ref{sec:benchmark_face}, except for training with $\pm90^{\circ}$ rotation augmentation and 1e-3 base learning rate.

\subsubsection{Cross-dataset Evaluation}
We conduct experiments to evaluate the cross-dataset generalization ability of WBH to unseen data.

\textbf{Benchmark Datasets.} We use three benchmark datasets. (1) Panoptic~\cite{simon2017hand} dataset is a large-scale dataset containing images captured in the CMU’s Panoptic studio with multi-view settings. Following~\cite{simon2017hand}, the training set of MPII+NZSL with manual keypoint annotations are also included for training. The evaluation is performed on the testing set of MPII+NZSL. (2) FreiHand~\cite{Freihand2019} dataset is a multi-view hand dataset recorded in front of a green screen and augmented with synthetic backgrounds. (3) One-Hand10K~\cite{Yangang2018Mask} is a recently proposed in-the-wild 2D hand pose estimation dataset with 10K annotated images. All of these datasets use 21-keypoint annotations, which is the same as WBH.

We use HRNetV2-W18~\cite{wang2020deep} as the base network and follow the same experimental settings as Sec.~\ref{sec:benchmark_hand} for training. Table~\ref{tab:hand_cross} summarizes the cross-dataset evaluation results. We choose one dataset for training, and the others for testing. It is expected that the model performs the best when the training and testing are conducted on the same dataset, and that the performance drops when the model is evaluated on unseen data. We observe that the model trained on FreiHand dataset~\cite{Freihand2019} do not generalize very well to other datasets. Because FreiHand is captured by 32 people against a green screen augmented with synthetic backgrounds. The diversity of the subjects is limited. We show that the model trained on WBH achieves reasonablely good performance on other datasets. Especially on Panoptic dataset, it surprisingly outperforms the model trained on Panoptic dataset (73.6\% vs 72.7\% AUC). This is because Panoptic dataset is captured in a controlled lab environment, while its test set (MPII+NZSL) is captured in the wild. We conclude that our proposed WBH achieves superior generalization capacity. This is because it consists of a large number of in-the-wild hand images with sufficient variation (\eg hand pose, view point, background, occlusion, and hand-object/hand-hand interaction).

\subsection{Ablation Study}
\label{sec:ablation_study}

\begin{table}[t]
	\caption{Effect of searching connections between BodyNet and FaceHead/HandHead on COCO-WholeBody V1.0 dataset. ``WB'' means ``whole-body''.}
	\label{tab:connection}
	\begin{center}
		\begin{tabular}{c|ccccc}
			\hline
			\multirow{2}{*}{Search Space} & body & foot & face & hand & WB \\
			~ & AP & AP & AP & AP & AP    \\
			\hline
			w/o Connection & 73.9 & 59.2 & 84.6 & 58.3 & 63.9 \\
			ZoomNAS  & 74.0 & 61.7 & 88.9 & 62.5 & 65.4  \\
			\hline
		\end{tabular}
	\end{center}
\end{table}

\begin{table*}[t]
	\caption{Effect of automatic computation allocation. Our proposed ZoomNAS shows the superiority over the search strategy of proportional computational complexity allocation. ``WB'' means ``whole-body''.}
	\label{tab:allocation}
	\begin{center}
		\begin{tabular}{c|ccca|cccca}
			\hline
			\multirow{2}{*}{Method} & BodyNet & FaceHead & HandHead & Overall & body & foot & face & hand & WB \\
			~ & GFLOPs & GFLOPs & GFLOPs & GFLOPs & AP & AP & AP & AP & AP    \\  \hline
			ZoomNet & 16.02 & 4.19 & 4.14 $\times$ 2 & 28.49 & 74.5 & 60.9 & 88.0 & 57.9 & 63.0 \\
			Proportional Allocation & 10.20 & 2.65 & 2.62 $\times$ 2 & 18.10 & 70.0 & 45.2 & 91.3 & 62.4 & 62.0 \\
			Automatic Allocation (ZoomNAS) & 12.71 & 0.57 & 2.37 $\times$ 2 & 18.02 & 74.0 & 61.7 & 88.9 & 62.5 & 65.4 \\
			\hline
		\end{tabular}
	\end{center}
\end{table*}

\textbf{Effect of connection search.} 
We search the connections between BodyNet and FaceHead/HandHead on two search spaces, \ie feature stage and RoI expansion. We enforce the baseline model with the same connections as ZoomNet (feature stage 0 and RoI expansion 1.0) and evaluate the model performance as ``w/o Connection'' in Table~\ref{tab:connection}. We show that searching for appropriate connections significantly improves the performance of whole-body pose estimation (65.2\% vs 63.9\% mAP), especially for face and hand keypoints.

\textbf{Effect of automatic computation allocation.}
As shown in Table~\ref{tab:allocation}, we investigate the effect of automatic computation allocation among BodyNet/FaceHead/HandHead. We search for a model with the same Flops as ZoomNAS and the computational complexity of BodyNet/FaceHead/ HandHead is reduced in equal proportions. In comparison, we find that ZoomNAS allocates more computational complexity for BodyNet. This is because BodyNet estimates keypoints of face/foot as well as provides features and boxes for FaceHead/HandHead, which is critical for whole-body performance. As a result, ZoomNAS with automatic computation allocation achieves higher accuracy (65.4\% vs 62.0\% mAP), which demonstrates the superiority of our proposed search strategy.

\begin{table}[t]
	\caption{Effect of training strategy. Comparing with optimizing BodyNet/FaceHead/HandHead successively, the joint training strategy achieves the comparable performance with much fewer training iterations.}
	\label{tab:training}
	\begin{center}
		\begin{tabular}{c|ccccc}
			\hline
			Training & body & foot & face & hand & WB \\ 
			Strategy & AP & AP & AP & AP & AP    \\
			\hline
			Successive Training & 73.9 & 59.2 & 89.3 & 60.8 & 65.2 \\
			Joint Training (ZoomNAS)  & 74.0 & 61.7 & 88.9 & 62.5 & 65.4  \\
			\hline
		\end{tabular}
	\end{center}
\end{table}

\textbf{Effect of joint training.} \label{sec:effect of joint training}
To explore the effect of joint training strategy, we build a baseline that optimizes BodyNet/FaceHead/HandHead successively, and each module is trained for 210 epochs. In comparison, these sub-modules in ZoomNAS is trained as a whole as introduced in Section~\ref{sec:train_zoomnet}, which simplifies and speeds up the training process. We compare the proposed joint training strategy with the successive training baseline in Table~\ref{tab:training}. The results show that the joint training strategy achieves comparable performance with complicated separate training, which validates the effectiveness of joint training.

\begin{table}[t]
	\caption{The improvement of COCO-WholeBody dataset. COCO-WholeBody V1.0 improves the annotations of training set and uses the same validation set as V0.5~\cite{jin2020whole}.}
	\label{tab:data}
	\begin{center}
		\begin{tabular}{c|c|ccccc}
			\hline
			\multirow{2}{*}{Data} & \multirow{2}{*}{Method} & body & foot & face & hand & WB \\
			~ & ~ & AP & AP & AP & AP & AP    \\
			\hline
			\multirow{4}{*}{V0.5~\cite{jin2020whole}} & OpenPose~\cite{cao2018openpose} & 56.3 & 53.2 & 48.2 & 19.8 & 33.8 \\
			~ & AE~\cite{newell2017associative} & 40.5 & 7.7 & 47.7 & 34.1 & 27.4 \\
			~ & HRNet~\cite{sun2019deep}  & 65.9 & 31.4 & 52.3 & 30.0 & 43.2 \\
			~ & ZoomNet & 74.3 & 79.8 & 62.3 & 40.1 & 54.1 \\ \hline
			\multirow{4}{*}{V1.0} & OpenPose~\cite{cao2018openpose} & 56.3 & 53.2 & 76.5 & 38.6 & 44.2 \\
			~ & AE~\cite{newell2017associative} & 58.0 & 57.7 & 58.8 & 48.1 & 44.0  \\
			~ & HRNet~\cite{sun2019deep}  & 70.1 & 58.6 & 72.7 & 51.6 & 58.6 \\
			~ & ZoomNet & 74.5 & 60.9 & 88.0 & 57.9 & 63.0 \\ \hline
		\end{tabular}
	\end{center}
\end{table}

\textbf{Effect of the improvement of COCO-WholeBody V1.0.}
As introduced in Section~\ref{sec:data_v1.0}, we improve both quantity and quality of the training set of COCO-Wholebody from V0.5~\cite{jin2020whole} to V1.0. As the validation set stays the same, we compare the baseline methods trained on the data of two versions. As shown in Table~\ref{tab:data}, all the approaches (\ie OpenPose~\cite{cao2018openpose}, AE~\cite{newell2017associative}, HRNet~\cite{sun2019deep}, and ZoomNet) achieve significantly better performance. For example, OpenPose achieves 10.4\% mAP improvement (44.2\% vs 33.8\% whole-body AP) with more accurate results of face (76.5\% vs 48.2\% face AP) and hand (38.6\% vs 19.8\% hand AP), which demonstrates the advance of COCO-Wholebody V1.0.

\section{Conclusion}
\label{sec:conclusion}
In this paper, we propose a top-down 2D whole-body pose estimation approach, termed ZoomNet. It takes into account the hierarchical structure of the full human body, and effectively addresses the problem of scale variance. We further propose ZoomNAS to jointly search the architectures and the connections of different sub-modules via neural architecture search. The computational complexity is automatically allocated to discovered sub-modules for the global optimum.
To meet the need for a whole-body pose estimation dataset, we propose COCO-WholeBody V1.0 with 200K+ high-quality whole-body annotations. 
Extensive experiments show the superiority of our proposed approaches and the significance of our introduced datasets.

\ifCLASSOPTIONcompsoc
  \section*{Acknowledgments}
  
\else
  \section*{Acknowledgment}
\fi

This work is supported in part by the General Research Fund through the Research Grants Council of Hong Kong under Grants (Nos. 14202217, 14203118, 14208619), in part by Research Impact Fund Grant No. R5001-18. Wanli Ouyang is supported by the Australian Research Council Grant DP200103223, FT210100228, and Australian Medical Research Future Fund MRFAI000085, CRC-P ``ARIA - Bionic Visual-Spatial Prosthesis for the Blin'' and ``Smart Material Recovery Facility (SMRF) - Curby Soft Plastic'', and SenseTime. Ping Luo is partially supported by the General Research Fund of Hong Kong No.27208720.


\ifCLASSOPTIONcaptionsoff
  \newpage
\fi


\bibliographystyle{IEEEtran}
\bibliography{bib/ref,bib/face,bib/hand,bib/nas,bib/dataset}

\vspace{-5pt}
\begin{IEEEbiography}[{\includegraphics[width=1in,height=1.25in,clip,keepaspectratio]{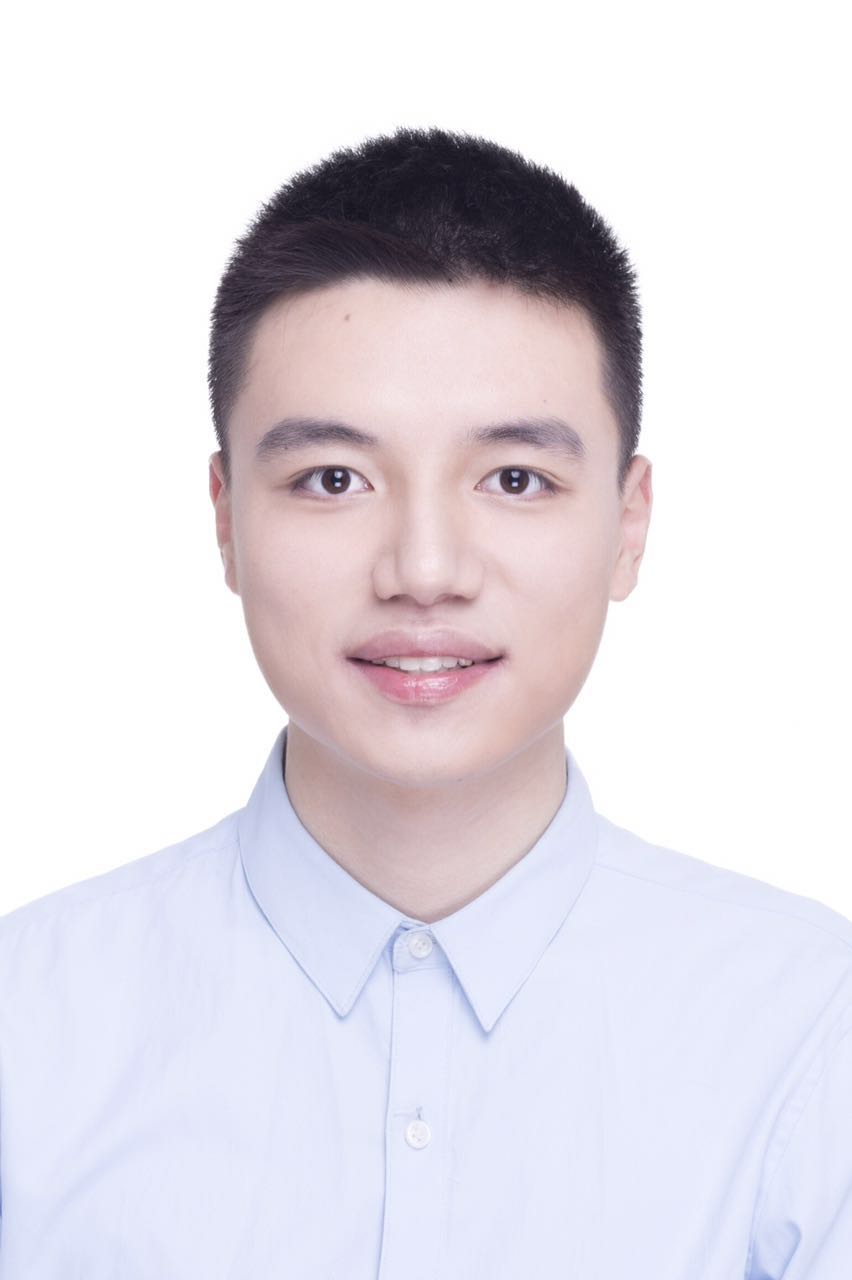}}]{Lumin Xu}
received the B.Eng. degree in information engineering from Zhejiang University, Hangzhou, China, in 2018. He is currently a Ph.D. candidate with the Department of Electronic Engineering, The Chinese University of Hong Kong, Hong Kong SAR, China. His research interests include computer vision, deep learning, and human pose estimation.
\end{IEEEbiography}

\vspace{-5pt}
\begin{IEEEbiography}[{\includegraphics[width=1in,height=1.25in,clip,keepaspectratio]{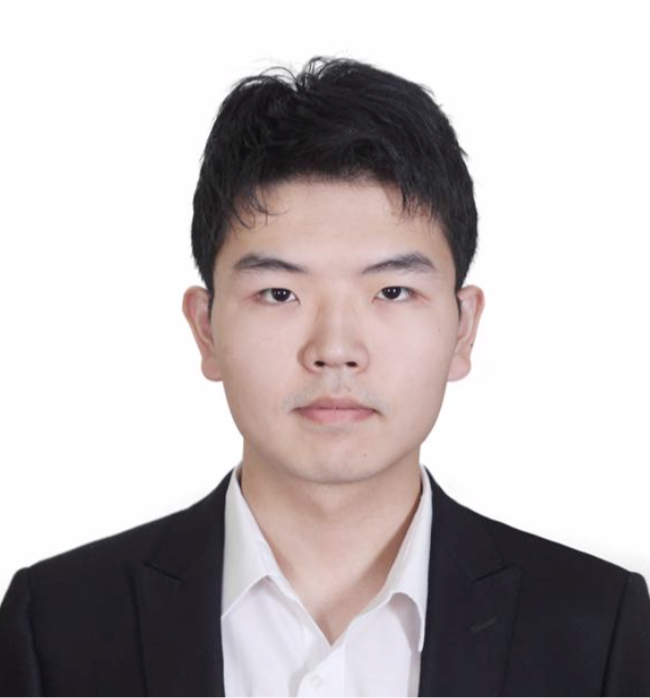}}]{Sheng Jin}
received the B.Eng. and M.Eng. degrees from the Department of Automation, Tsinghua University, Beijing, China, in 2017 and 2020. He is currently a Ph.D. student at the University of Hong Kong, Hong Kong SAR, China. His research interests include deep learning and human pose estimation.
\end{IEEEbiography}

\vspace{-5pt}
\begin{IEEEbiography}[{\includegraphics[width=1in,height=1.25in,clip,keepaspectratio]{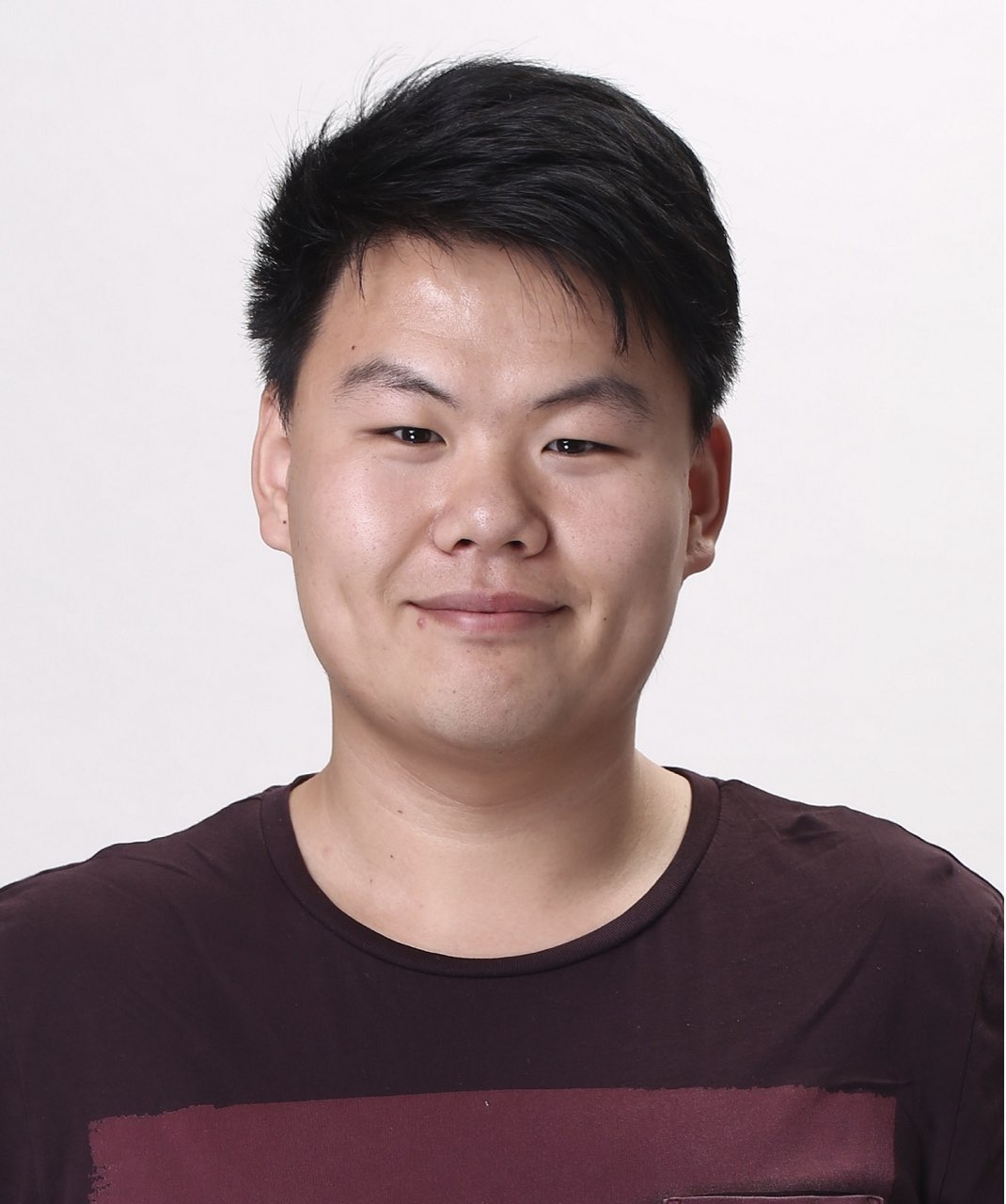}}]{Wentao Liu} received his Ph.D. degree in the School of EECS, Peking University. He is currently the Research Director of SenseTime, responsible for end-edge computing research. The research products are widely applied in augmented reality, smart industry, and business intelligence. His research interests include computer vision and pattern recognition.
\end{IEEEbiography}

\begin{IEEEbiography}[{\includegraphics[width=1in,height=1.25in,clip,keepaspectratio]{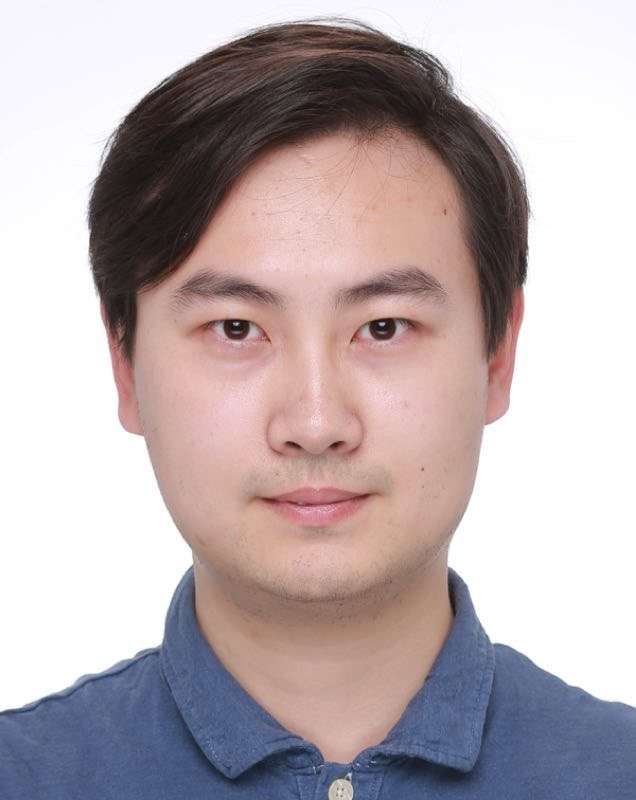}}]{Chen Qian} is currently the Executive Research Director of SenseTime, where he is responsible for leading the team in AI content generation and end-edge computing research in 2D and 3D scenarios. The technology is widely used in the top four mobile companies in China, APPs both home and abroad in augmented reality, video sharing and live streaming, vehicle OEMs, and smart industry. He has published dozens of articles on top journals and dozens of papers on top conferences, such as TPAMI, CVPR, ICCV, and ECCV with more than 4000 citations. He has also led the team to achieve the first place in the Competition of Face Identification and Face Verification in Megaface Challenge.
\end{IEEEbiography}

\begin{IEEEbiography}[{\includegraphics[width=1in,height=1.25in,clip,keepaspectratio]{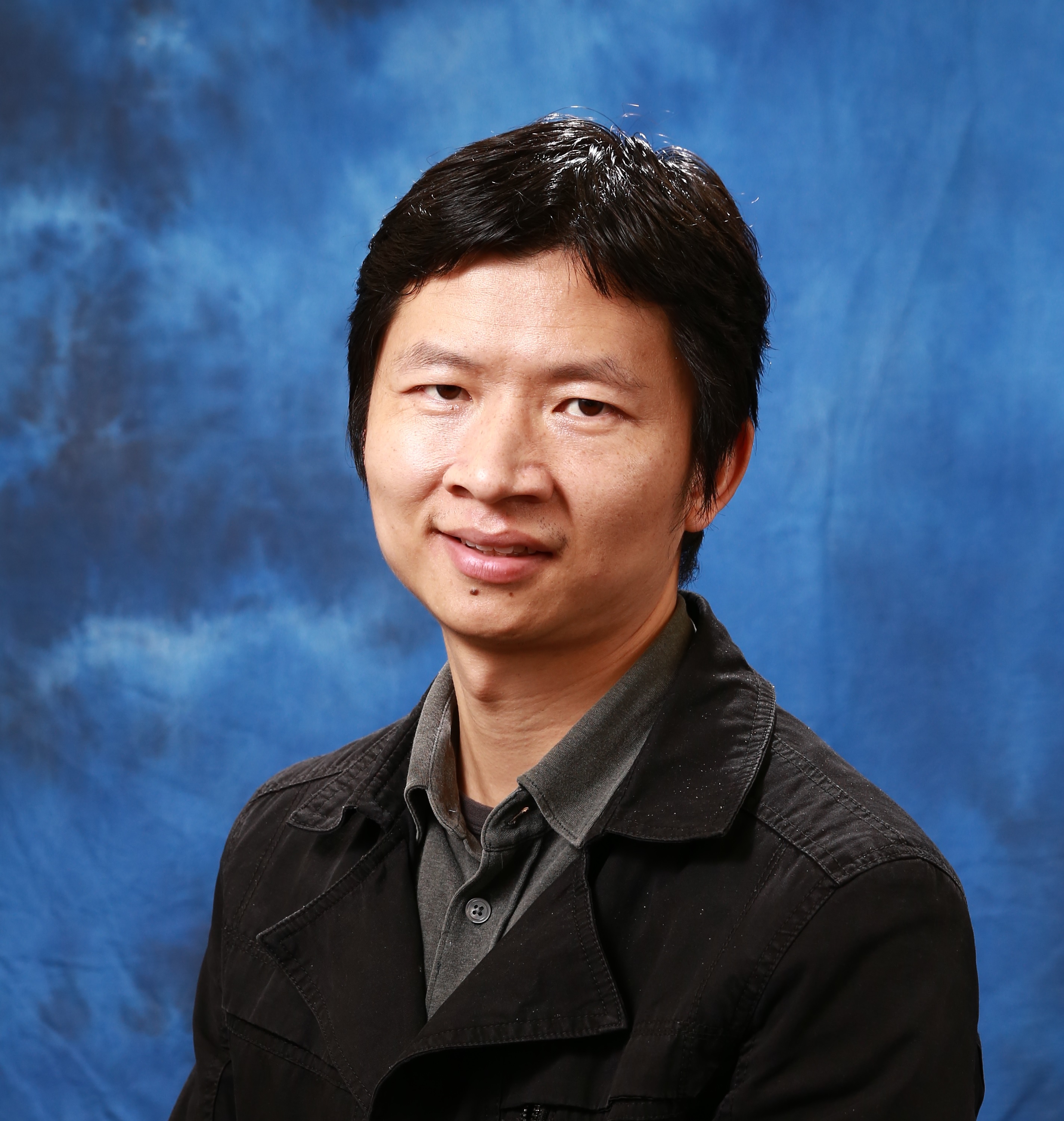}}]
{Wanli Ouyang} received the PhD degree in the Department of Electronic Engineering, The Chinese University of Hong Kong. He is now an associate professor in the School of Electrical and Information
Engineering at the University of Sydney, Australia. His research interests include image processing, computer vision and pattern recognition. He is a senior member of IEEE.
\end{IEEEbiography}

\begin{IEEEbiography}[{\includegraphics[width=1in,height=1.25in,clip,keepaspectratio]{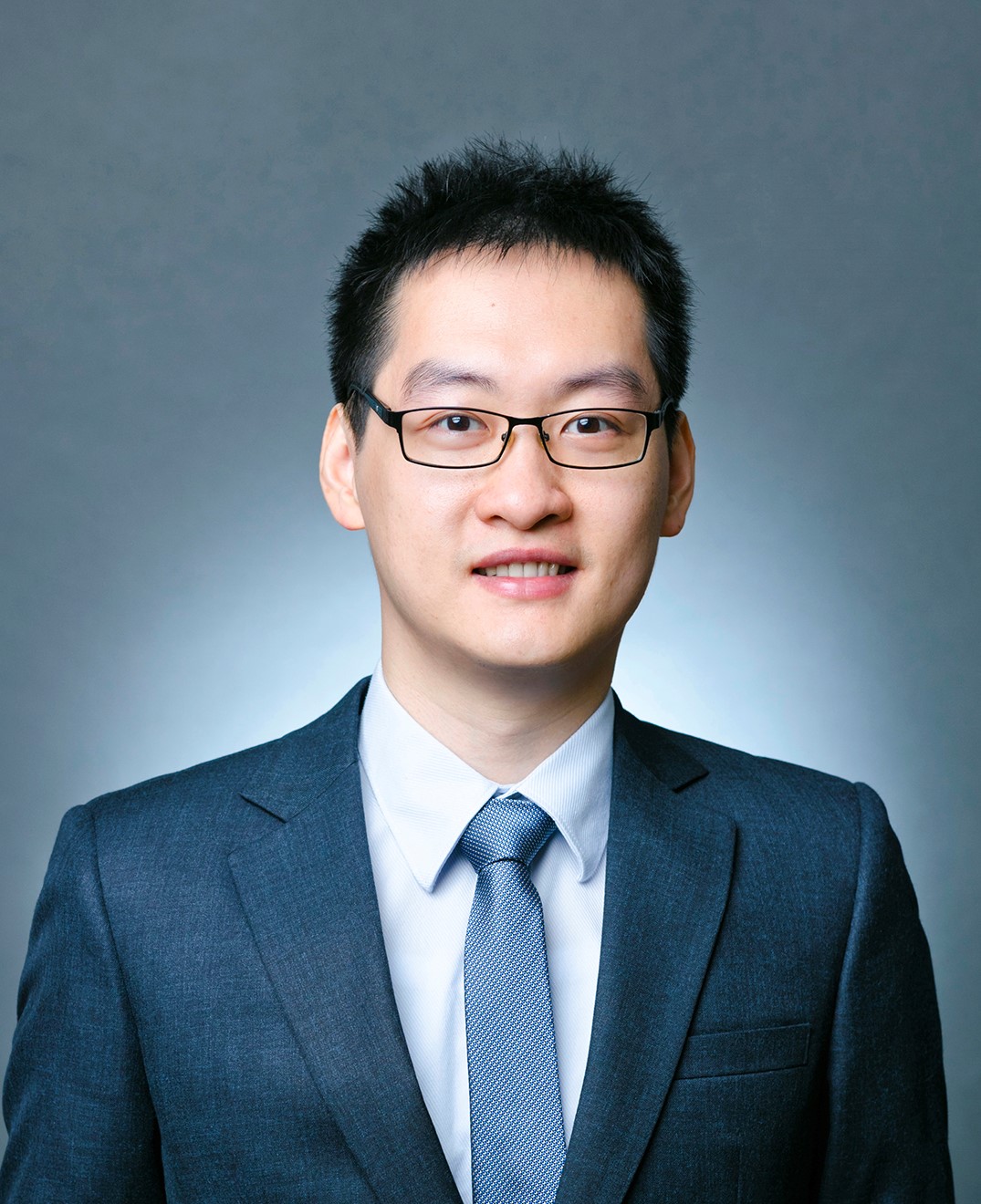}}]{Ping Luo}
is an Assistant Professor in the department of computer science, The University of Hong Kong (HKU). 
He received his PhD degree in 2014 from Information Engineering, the Chinese University of Hong Kong (CUHK), supervised by Prof. Xiaoou Tang and Prof. Xiaogang Wang. He was a Postdoctoral Fellow in CUHK from 2014 to 2016. 
He joined SenseTime Research as a Principal Research Scientist from 2017 to 2018. 
His research interests are machine learning and computer vision. He has published 100+ peer-reviewed articles in top-tier conferences and journals such as TPAMI, IJCV, ICML, ICLR, CVPR, and NIPS. His work has high impact with 18000+ citations according to Google Scholar. He has won a number of competitions and awards such as the first runner up in 2014 ImageNet ILSVRC Challenge, the first place in 2017 DAVIS Challenge on Video Object Segmentation, Gold medal in 2017 Youtube 8M Video Classification Challenge, the first place in 2018 Drivable Area Segmentation Challenge for Autonomous Driving, 2011 HK PhD Fellow Award, and 2013 Microsoft Research Fellow Award (ten PhDs in Asia).
\end{IEEEbiography}

\begin{IEEEbiography}[{\includegraphics[width=1in,height=1.25in]{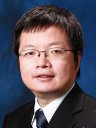}}]{Xiaogang Wang}
received the B.S. degree from the University of Science and Technology of China in 2001, the MS degree from The Chinese University of Hong Kong in 2003, and the PhD degree from the Computer Science and Artificial Intelligence Laboratory, Massachusetts Institute of Technology in 2009. He is currently a professor in the Department of Electronic Engineering at The Chinese University of Hong Kong. His research interests include computer vision and machine learning. 
\end{IEEEbiography}










\section*{Supplementary material}

# APPENDIX A
# MORE EXPERIMENTS ON 2D FACE KEYPOINT LOCALIZATION

To further explore the generalization ability of our proposed WholeBody-Face (WBF) dataset for face keypoint localization, we perform cross-dataset experiments on four benchmark datasets including AFLW [1], COFW [2], WFLW [3], and 300W [4].

**AFLW** [1] dataset consists of 20,000 training images with 19 annotated facial landmarks. The evaluation is conducted on the AFLW-Full set (4,386 testing images) and the AFLW-Frontal set (1,314 testing images).

**COFW** [2] dataset consists of 1,345 training and 507 testing images, where each face is annotated with 29 facial landmarks.

**WFLW** [3] dataset is built based on the WIDER Face dataset [5]. WFLW dataset consists of 7,500 training and 2,500 testing images with 98 manually annotated landmarks. The evaluation is conducted on the test set and several subsets: large pose (326 images), expression (314 images), illumination (698 images), make-up (206 images), occlusion (736 images), and blur (773 images).

**300W** [4] dataset is a combination of LFPW [6], AFW [7], HELEN [8], and XM2VTS [9] datasets. It contains 3,148 training images, where each face has 68 annotated landmarks. We follow the common settings [10] to train models on the training images, validate on the "common" set (554 images), and evaluate on the "challenging" (135 images), "full" (689 images), and "test" (300 indoor and 300 outdoor images) sets.

TABLE 1: Cross-dataset evalutation on WBF dataset. We pre-train HRNetV2-W18 [10] on benchmark datasets and test on WBF dataset. "A $\Rightarrow$ B" means pre-training on dataset A and fine-tuning on dataset B. $\downarrow$ means lower is better.

| Train Set | NME $\downarrow$ |
|---|---|
| WBF | 5.68 |
| AFLW $\Rightarrow$ WBF | 5.25 |
| COFW $\Rightarrow$ WBF | 5.22 |
| WFLW $\Rightarrow$ WBF | 5.71 |
| 300W $\Rightarrow$ WBF | 5.44 |

We experiment by pre-training on these four benchmark datasets and testing on the WBF dataset in Table 1. HRNetV2-W18 [10] is applied for face landmark detection, which is the current state-of-the-art model. We use the normalized mean error (NME) for evaluation with interocular distance as normalization. We find that pre-training on AFLW, 300W, and COFW improves the performance on WBF. Interestingly, we also notice that pre-training on WFLW does not bring any improvement. This is partially due to that WBF is sufficiently large enough to train the model well and that there are large domain gaps between WFLW and WBF.

We further show that our proposed WBF dataset is complementary to other benchmark datasets, by pre-training the model on WBF dataset. HRNetV2-W18 [10] is adopted as the facial landmark detection model and the normalized mean error (NME) is used for evaluation.

Table 2 presents the facial keypoint estimation results on AFLW dataset. The performance of HRNetV2-W18 improves by pre-training on WBF, which even outperforms DCFE [23] with extra 3D information and PDB [24] with stronger data augmentation. The model pre-trained on WBF only performs slightly worse than LAB [3] which uses extra boundary information.

TABLE 2: Facial keypoint estimation results on AFLW. "WBF $\Rightarrow$" means pre-training on WBF dataset, which improves upon the state-of-the-art HRNetV2 model and even outperforms several models trained with extra information. "w/ 3D" means using extra 3D information, "w/ DA" means utilizing stronger data augmentation, and "w/ B" means relying on extra boundary information. NMS is adopted for evaluation and $\downarrow$ means lower is better.

| Method | Backbone | Full $\downarrow$ | Frontal $\downarrow$ |
|---|---|---|---|
| RCN [11] | - | 5.60 | 5.36 |
| CDM [12] | - | 5.43 | 3.77 |
| ERT [13] | - | 4.35 | 2.75 |
| LBF [14] | - | 4.25 | 2.74 |
| SDM [15] | - | 4.05 | 2.94 |
| CFSS [16] | - | 3.92 | 2.68 |
| RCPR [2] | - | 3.73 | 2.87 |
| CCL [17] | - | 2.72 | 2.17 |
| DAC-CSR [18] | - | 2.27 | 1.81 |
| TSR [19] | VGG-S | 2.17 | - |
| CPM + SBR [20] | CPM | 2.14 | - |
| SAN [21] | ResNet-152 | 1.91 | 1.85 |
| DSRN [22] | - | 1.86 | - |
| LAB (w/o B) [3] | Hourglass | 1.85 | 1.62 |
| HRNetV2 [10] | HRNetV2-W18 | 1.57 | 1.46 |
| WBF $\Rightarrow$ HRNetV2 [10] | HRNetV2-W18 | **1.42** | **1.27** |
| *Model trained with extra info.* | | | |
| DCFE (w/ 3D) [23] | - | 2.17 | - |
| PDB (w/ DA) [24] | ResNet-50 | 1.47 | - |
| LAB (w/ B) [3] | Hourglass | 1.25 | 1.14 |

TABLE 3: Facial keypoint estimation results on COFW. NME and $FR_{0.1}$ are adopted for evaluation. $FR_{0.1}$ measures the failure rate at the normalized threshold 0.1. $\downarrow$ means lower is better.

| Method | Backbone | NME $\downarrow$ | $FR_{0.1}$ $\downarrow$ |
|---|---|---|---|
| ESR [25] | - | 11.20 | 36.00 |
| RCPR [2] | - | 8.50 | 20.00 |
| HPM [26] | - | 7.50 | 13.00 |
| CCR [27] | - | 7.03 | 10.90 |
| DRDA [28] | - | 6.46 | 6.00 |
| RAR [29] | - | 6.03 | 4.14 |
| DAC-CSR [18] | - | 6.03 | 4.73 |
| LAB (w/o B) [3] | Hourglass | 5.58 | 2.76 |
| HRNetV2 [10] | HRNetV2-W18 | 3.45 | **0.19** |
| WBF $\Rightarrow$ HRNetV2 [10] | HRNetV2-W18 | **3.36** | **0.19** |
| *Model trained with extra info.* | | | |
| PDB (w/ DA) [24] | ResNet-50 | 5.07 | 3.16 |
| LAB (w/ B) [3] | Hourglass | 3.92 | 0.39 |

Table 3 provides the facial keypoint localization results on COFW. We adopt NME and $FR_{0.1}$ to evaluate the performance of all the methods. $FR_{0.1}$ measures the failure rate, where the distance between the predicted and ground-truth keypoint locations is larger than the normalized threshold of 0.1. HRNetV2-W18 pre-trained on WBF significantly outperforms the existing facial landmark detection approaches, including those utilizing extra information.

Table 4 provides the comparisons on WFLW dataset. The performance of facial landmark detection is evaluated on both the test set and several subsets. Pre-training on WBF significantly promotes the accuracy of keypoint localization

TABLE 4: Facial landmark detection results on WFLW test set and 6 subsets including pose, expression (expr.), illumination (illu.), make-up (mu.), occlusion (occu.) and blur. Pre-training on WBF achieves significantly better performance on all the test sets than the prior arts, especially on these challenging subsets (pose, occu., and blur).

| Method | Backbone | test ↓ | pose ↓ | expr. ↓ | illu. ↓ | mu ↓ | occu. ↓ | blur ↓ |
|---|---|---|---|---|---|---|---|---|
| ESR [25] | - | 11.13 | 25.88 | 11.47 | 10.49 | 11.05 | 13.75 | 12.20 |
| SDM [15] | - | 10.29 | 24.10 | 11.45 | 9.32 | 9.38 | 13.03 | 11.28 |
| CFSS [16] | - | 9.07 | 21.36 | 10.09 | 8.30 | 8.74 | 11.76 | 9.96 |
| DVLN [30] | VGG-16 | 6.08 | 11.54 | 6.78 | 5.73 | 5.98 | 7.33 | 6.88 |
| HRNetV2 [10] | HRNetV2-W18 | 4.60 | 7.94 | 4.85 | 4.55 | 4.29 | 5.44 | 5.42 |
| WBF ⇒ HRNetV2 [10] | HRNetV2-W18 | **4.01** | **6.56** | **4.32** | **3.93** | **3.89** | **4.73** | **4.56** |
| *Model trained with extra info.* | | | | | | | | |
| LAB (w/ B) [3] | Hourglass | 5.27 | 10.24 | 5.51 | 5.23 | 5.15 | 6.79 | 6.32 |
| PDB (w/ DA) [24] | ResNet-50 | 5.11 | 8.75 | 5.36 | 4.93 | 5.41 | 6.37 | 5.81 |

TABLE 5: Facial keypoint estimation (NME) on 300W: "common" (for val), "challenging", "full" and "test". Pre-training on WBF improves the performance of HRNetV2 on the validation set and the test sets, especially on 300W-challenging.

| Method | Backbone | common ↓ | challenging ↓ | full ↓ | test ↓ |
|---|---|---|---|---|---|
| RCN [11] | - | 4.67 | 8.44 | 5.41 | - |
| DSRN [22] | - | 4.12 | 9.68 | 5.21 | - |
| PCD-CNN [31] | - | 3.67 | 7.62 | 4.44 | - |
| CPM + SBR [20] | CPM | 3.28 | 7.58 | 4.10 | - |
| SAN [21] | ResNet-152 | 3.34 | 6.60 | 3.98 | |
| DAN [32] | - | 3.19 | 5.24 | 3.59 | 4.30 |
| HRNetV2 [10] | HRNetV2-W18 | 2.87 | 5.15 | 3.32 | 3.85 |
| WBF ⇒ HRNetV2 [10] | HRNetV2-W18 | **2.84** | **4.73** | **3.21** | **3.68** |
| *Model trained with extra info.* | | | | | |
| LAB (w/ B) [3] | Hourglass | 2.98 | 5.19 | 3.49 | - |
| DCFE (w/ 3D) [23] | - | 2.76 | 5.22 | 3.24 | 3.88 |

and performs better than all the other methods on the test set and the subsets. Especially, the improvement is remarkable on pose (6.56 vs 7.94), occlusion (4.73 vs 5.44) and blur (4.56 vs 5.42) subsets, as WBF has a large variance of poses and contains many challenging images.

Table 5 shows the performance of facial keypoint estimation on 300W dataset. Pre-training on WBF achieves significantly better performance on "challenging", "full", and "test" sets than the prior arts. On 300W-challenging, the performance improvement is up to 8% (4.73 vs 5.15), which demonstrates that our proposed large-scale in-the-wild WBF is helpful for cases with complex backgrounds, pose, and expression variations.

In conclusion, we show that pre-training on WBF dataset improves the performance of the state-of-the-art model (HRNetV2) and outperforms the existing facial keypoint localization methods by a large margin. Our proposed WBF dataset is complementary to benchmark datasets with strong generalization ability, and can promote the development of facial landmark detection.

# APPENDIX B
# MORE EXPERIMENTS ON 2D HAND KEYPOINT LOCALIZATION

To investigate the effect of our proposed 2D hand keypoint localization dataset, WholeBody-Hand(WBH), we perform cross-dataset experiments on four benchmark datasets: Panoptic [33], FreiHand [34], RHD [35], and OneHand10K [36].

**Panoptic** [33] dataset is a widely used hand pose estimation dataset. The images are captured in a controlled lab environment, *i.e.* the CMU's Panoptic studio with multi-view dome settings. Following [33], a small set of images from MPII+NZSL with manual keypoint annotations are also used for training (1,912 hands), resulting in a total of 16K training annotations. The evaluation is performed on the testing set of MPII+NZSL (846 hands).

**FreiHand** [34] dataset is a popular multi-view hand dataset, recorded in front of a green screen and augmented with synthetic backgrounds. The dataset is recorded by 32 actors. It contains 32,560 images for training and 3,960 images for evaluation.

**Rendered Hand Pose (RHD)** [35] dataset is a synthetic dataset built upon 20 characters. In total, it contains 41,258 images for training and 2,728 images for evaluation.

**One-Hand10K** [36] is a recently proposed in-the-wild 2D hand pose estimation dataset. It contains 10,000 images for training and 1,703 images for evaluation.

TABLE 6: Cross-dataset evaluation for WBH dataset. HRNetV2-W18 is adopted for hand keypoint localization. "A ⇒ B" means pre-training on dataset A and fine-tuning on dataset B. ↑ means higher is better, while ↓ means lower is better.

| Train Set | Test Set | PCK ↑ | AUC ↑ | EPE ↓ |
|---|---|---|---|---|
| Panoptic | Panoptic | 0.998 | 0.727 | 8.46 |
| WBH ⇒ Panoptic | Panoptic | 0.999 | 0.775 | 6.44 |
| OneHand10K | OneHand10K | 0.987 | 0.556 | 26.7 |
| WBH ⇒ OneHand10K | OneHand10K | 0.990 | 0.559 | 25.8 |
| FreiHand | FreiHand | 0.993 | 0.869 | 3.21 |
| WBH ⇒ FreiHand | FreiHand | 0.994 | 0.870 | 3.17 |
| RHD | RHD | 0.988 | 0.895 | 2.41 |
| WBH ⇒ RHD | RHD | 0.991 | 0.900 | 2.27 |
| WBH | WBH | 0.802 | 0.834 | 4.57 |
| Panoptic ⇒ WBH | WBH | 0.804 | 0.837 | 4.50 |
| OneHand10K ⇒ WBH | WBH | 0.804 | 0.836 | 4.52 |
| RHD ⇒ WBH | WBH | 0.806 | 0.837 | 4.49 |

In Table 6, we analyze the generalization ability of

TABLE 7: Quantitative performance results on 2D hand pose estimation. "WBH ⇒" means pre-training on WBH. ↑ means higher is better, while ↓ means lower is better.

| Method | PCK ↑ | AUC ↑ | EPE ↓ |
|---|---|---|---|
| Panoptic Dataset | | | |
| Zimmermann et al. [35] | - | 0.170 | 59.4 |
| Boukhayma et al [37] | - | 0.500 | 18.9 |
| Santavas et al [38] | - | 0.550 | 16.1 |
| HRNetV2 [10] | 0.998 | 0.727 | 8.40 |
| WBH ⇒ HRNetV2 [10] | **0.999** | **0.775** | **6.44** |
| FreiHand Dataset | | | |
| Santavas et al [38] | - | 0.870 | 4.00 |
| SBL-res50 [39] | 0.992 | 0.868 | 3.27 |
| HRNetV2 [10] | 0.993 | 0.869 | 3.21 |
| WBH ⇒ HRNetV2 [10] | **0.994** | **0.870** | **3.17** |
| RHD Dataset | | | |
| Zimmermann. et al. [35] | 0.800 | 0.724 | 9.14 |
| HRNetV2 [10] | 0.988 | 0.895 | 2.41 |
| WBH ⇒ HRNetV2 [10] | **0.991** | **0.900** | **2.27** |
| OneHand10K Dataset | | | |
| CPM [40] | 0.885 | - | - |
| LPM-G1 [41] | 0.896 | - | - |
| RealMask [36] | 0.898 | - | - |
| HRNetV2 [10] | 0.987 | 0.556 | 26.7 |
| WBH ⇒ HRNetV2 [10] | **0.990** | **0.559** | **25.8** |

WholeBody-Hand (WBH) by cross-dataset evaluation. We adopt HRNetV2-W18 for hand keypoint localization and report EPE, PCK, and AUC to evaluate the performance of 2D hand pose estimation. "A ⇒ B" means pre-training on dataset A and fine-tuning on dataset B. We notice that pre-training on WBH improves upon the baseline by a large margin. For example, on the Panoptic dataset, AUC improves from 0.727 to 0.775 and EPE decreases from 8.46 to 6.44. And pre-training on other datasets also helps to improve the performance on WBH. We conclude that our proposed WBH is complementary to other benchmark datasets. In Table 7, we compare with the existing hand pose estimation approaches on the benchmark datasets. The results show that we set the new state-of-the-arts by pre-training on WBH dataset.



\end{document}